%% file: main.tex
\documentclass{article}

\usepackage{microtype}
\usepackage{graphicx}
\usepackage{subfigure}
\usepackage{booktabs} 

\usepackage{hyperref}

\usepackage{balance}

\usepackage[preprint]{icml2024}

\usepackage{amsmath}
\usepackage{amssymb}
\usepackage{mathtools}
\usepackage{amsthm}

\usepackage[capitalize,noabbrev]{cleveref}

\theoremstyle{plain}

\theoremstyle{definition}

\theoremstyle{remark}

\usepackage{tikz}
\newcommand{\mtick}{\tikz\fill[scale=0.4](0,.35) -- (.25,0) -- (1,.7) -- (.25,.15) -- cycle;}
\newcommand{\mcross}{%
\tikz[scale=0.23] {
    \draw[line width=0.7,line cap=round] (0,0) to [bend left=6] (1,1);
    \draw[line width=0.7,line cap=round] (0.2,0.95) to [bend right=3] (0.8,0.05);
}}

\newcommand{\ssubsec}[1]{\noindent\textbf{#1}}
\newcommand{\rbs}[1]{\textcolor{red}{\textbf{#1}}}
\newcommand{\rgd}[1]{\textcolor{violet}{#1}}

\definecolor{myred}{rgb}{0.8, 0, 0}
\definecolor{mygrn}{rgb}{0, 0.4, 0}
\newcommand{\colred}[1]{\textcolor{myred}{#1}}
\newcommand{\colgrn}[1]{\textcolor{mygrn}{#1}}

\newcommand{\istd}[1]{\textsubscript{(#1)}}
\newcommand{\tcap}[1]{\textbf{#1}}
\newcommand{\lob}{{$\downarrow$}}
\newcommand{\hib}{{$\uparrow$}}

\newcommand{\tsup}[1]{$^\text{#1}$}

\newcommand{\tsupsub}[2]{$^\text{#1}_\text{#2}$}

\newcommand{\softmax}{{\mathrm{softmax}}}

\input{citations.tex}

\usepackage[textsize=tiny]{todonotes}

\graphicspath{{figures/}}
\icmltitlerunning{Triplet Interaction Improves Graph Transformers}

\begin{document}

\twocolumn[
\icmltitle{Triplet Interaction Improves Graph Transformers:\\
           Accurate Molecular Graph Learning with Triplet Graph Transformers}




\begin{icmlauthorlist}
\icmlauthor{Md Shamim Hussain}{inst}
\icmlauthor{Mohammed J. Zaki}{inst}
\icmlauthor{Dharmashankar Subramanian}{comp}
\end{icmlauthorlist}

\icmlaffiliation{inst}{Department of Computer Science, Rensselaer Polytechnic Institute, Troy, NY, USA}
\icmlaffiliation{comp}{IBM Thomas J. Watson Research Center, Yorktown Heights, NY, USA}

\icmlcorrespondingauthor{Md Shamim Hussain}{hussam4@rpi.edu}
\icmlcorrespondingauthor{Mohammed J. Zaki}{zaki@cs.rpi.edu}

\icmlkeywords{Graph Neural Networks, Graph Transformers, Molecular Graphs, Triplet Interaction}

\vskip 0.3in
]



\printAffiliationsAndNotice{}  


\begin{abstract}
    Graph transformers typically lack third-order interactions, limiting their geometric understanding which is crucial for tasks like molecular geometry prediction. We propose the Triplet Graph Transformer (TGT) that enables direct communication between pairs within a 3-tuple of nodes via novel triplet attention and aggregation mechanisms. TGT is applied to molecular property prediction by first predicting interatomic distances from 2D graphs and then using these distances for downstream tasks. A novel three-stage training procedure and stochastic inference further improve training efficiency and model performance. Our model achieves new state-of-the-art (SOTA) results on open challenge benchmarks PCQM4Mv2 and OC20 IS2RE. We also obtain SOTA results on QM9, MOLPCBA, and LIT-PCBA molecular property prediction benchmarks via transfer learning. We also demonstrate the generality of TGT with SOTA results on the traveling salesman problem (TSP).
\end{abstract}

\section{Introduction}
\label{sec:intro}
Recent works have demonstrated the effectiveness of transformer \citep{vaswani2017attention} architectures across various data modalities. Originally developed for textual data, the transformer has since been adapted to image \citep{dosovitskiy2020image} and audio \citep{child2019generating}, achieving state-of-the-art (SOTA) results. More recently, pure graph transformers (GTs) \citep{ying2021transformers,hussain2022global,park2022grpe} have emerged as a promising architecture for graph-structured data, outperforming prior approaches involving localized convolutional/message-passing graph neural networks (MPNNs). First applied to molecular graphs, GTs have shown superior performance on diverse graph datasets including super-pixel and citation networks, and have been used to solve problems like vehicle routing \citeEdgeVRP{} and the traveling salesman (TSP) problem \citeEGT{}. This success is attributed to the ability of GTs to model long-range dependencies between nodes, overcoming the limitations of localized architectures.

Graph transformers utilize global self-attention to enable dynamic information exchange among node representations. Additionally, since graph topology and edge representations are as crucial as node representations for many tasks, the Edge-augmented Graph Transformer (EGT) \citep{hussain2022global} introduces dedicated edge channels that are updated across layers, enabling new pairwise (i.e., both existing and non-existing edge) representations to emerge over consecutive layers. This explicit modeling of both node and edge embeddings benefits performance on both node-centric and pairwise/link prediction tasks. However, although EGT allows information flow between node and pair representations, it lacks direct communication between pairs. Instead, neighboring pairs can only exchange information via their common node, creating a bottleneck. This limits the expressivity of the model by allowing only 2nd-order interactions \citep{joshi2023expressive}. As shown by \citet{li2024distance} 3rd-order interactions in the form of direct communication between neighboring pairs, i.e., within a 3-tuple of nodes is important for understanding important geometric concepts like angles.

In particular, 3D molecular geometry plays a vital role in determining chemical properties. While molecular graphs represent atoms as nodes and bonds as edges (“2D” structure), the relative positions of atoms in 3D space crucially influence quantum mechanical attributes like orbital energies and dipole moments and also other properties like solubility and interactions with proteins. Accordingly, prior works \citep{stark20223d,liu2021spherical} have shown that incorporating 3D geometry significantly improves performance on molecular property prediction. Geometric GNNs like GemNet \citep{gasteiger2021gemnet} rely on input features derived from the ground truth 3D geometry to predict molecular properties. However, determining ground truth 3D geometries requires expensive quantum chemical simulations, presenting computational barriers for large-scale inference. In this work, we explicitly learn to predict the molecular geometry using only 2D topological information. Specifically, we train a model to predict interatomic distances, serving as geometric input to the downstream chemical property prediction task. This is a paradigm shift from the traditional approach of 3D to 2D transfer learning \citep{stark20223d} or relying on less accurate 3D geometry \citep{fang2021chemrl}.

We introduce a novel 3rd-order interaction mechanism in the form of triplet interactions that enable direct communication between neighboring pairwise representations. This improves the expressivity of popular 2nd-order interaction based graph transformers like EGT. Reliance on 3rd-order \emph{interactions} rather than \emph{features} (similar to \citep{li2024distance}) allows for (i) accurate prediction of geometry from scratch, i.e., without an initial estimate of 3D coordinates, and (ii) robustness to input geometric inaccuracies. Geometric GNNs lack these capabilities due to their direct reliance on input geometric features. We propose two forms of triplet interactions called triplet attention and triplet aggregation. We call this architecture \emph{Triplet Graph Transformer (TGT)}. We also demonstrate the effectiveness of triplet interaction in other geometric graph learning tasks, such as the traveling salesman problem (TSP) -- demonstrating its generality.

We introduce a two-stage model for molecular property prediction involving a distance predictor and a task predictor. Unlike previous approaches like UniMol+ \citep{lu2023highly}, our method eliminates the need for initial (e.g., RDKit \citeRDKit{}) 3D coordinates, and instead learns to predict interatomic distances from 2D molecular graphs. The distance predictor can be directly used for other molecular property prediction tasks, whereas the task predictor can be finetuned for related quantum chemical prediction tasks.

We also propose a three-stage training framework for the distance and task predictors, which significantly reduces the training time and improves the performance of the model. We also introduce a novel stochastic inference technique that further improves the model's performance and allows for non-iterative parallel inference and uncertainty estimation. Additionally, we introduce new methods for regularizing both the pairwise update and triplet interaction mechanisms. We also propose a locally smooth structural noising method and a binned distance prediction objective that makes the model robust to structural perturbations.

Through these contributions, our proposed TGT model achieves new state-of-the-art (leaderboard) results on the PCQM4Mv2 \citeOGBLSC{}, OC20 IS2RE \citeOCtwenty{}, and QM9 \citeQMnine{} quantum chemical datasets. We also demonstrate the transferability of our learned distance predictor by achieving SOTA results on the MOLPCBA \citeOGB{} and LIT-PCBA \citeLITPCBA{} benchmarks, which are molecular property prediction and drug discovery datasets, respectively. This showcases the ability of our trained distance predictor to act as an off-the-shelf pairwise feature extractor that can be utilized for new molecular graph learning tasks.

\section{Related Work}
\label{sec:related}

Some previous works like GraphTrans \citep{wu2021representing}, GSA \citep{wang2021global}, and GROVER \citep{rong2020self} and some new works like GPS \citeGPS{} and GPS++ \citeGPSPP{} have used global self-attention to boost the expressivity of GNNs in a hybrid approach. But our work is more directly related to the recently proposed \emph{pure} graph transformers (GTs) such as SAN \citeSAN{}, Graphormer \citeGraphormer{}, EGT \citeEGT{}, GRPE \citeGRPE{}, GEM-2 \citeGEMtwo{}, and UniMol+ \citeUniMolP{}. Our contribution is introducing novel 3rd-order interaction mechanisms that improve the expressivity of GTs. We primarily approach the problem of molecular property prediction from a pure graph transformer perspective (but also demonstrate its generality). Recently, this problem has seen a lot of interest in the form of equivariant/invariant geometric GNNs like SchNet \citep{schutt2017schnet}, DimeNet \citep{gasteiger2020directional}, GemNet \citep{gasteiger2021gemnet}, SphereNet \citep{liu2021spherical} and equivariant transformers like Equiformer \citep{tholke2021equivariant}, and TorchMD-Net \citep{tholke2022torchmd}. Unlike these works, our model can be used for both general-purpose graph representation learning and geometric deep learning \citep{li2024distance}. We preserve SE(3) invariance by limiting the input features to interatomic distances. We train our network to predict interatomic distances from 2D molecular graphs, which allows for inference even in the absence of 3D information. This is in contrast to 3D pretraining approaches like 3D Infomax \citep{stark20223d}, GraphMVP \citep{liu2021pre}, Chemrl-GEM \citep{fang2021chemrl}, 3D PGT \citep{wang2023automated}, GeomSSL \citep{liu2022molecular}, Transformer-M \citep{luo2022one} which resort to 2D finetuning or multitask learning to make predictions in absence of 3D information. On the other hand, UniMol+ \citep{lu2023highly} iteratively refines cheaply computed RDKit \citeRDKit{} coordinates. In contrast, our approach requires no initial 3D coordinates and directly predicts interatomic distances from 2D graphs.

\section{Method}
\label{sec:method}

\subsection{TGT Architecture}
Triplet Graph Transformer (TGT) significantly extends the Edge-augmented Graph Transformer (EGT) \citep{hussain2022global} by introducing direct pair-to-pair communication in the form of triplet (3rd-order) interactions. In each layer, EGT maintains both node embedding $\mathbf{h}_i$ for each of the $N$ nodes and pairwise embedding $\mathbf{e}_{ij}$ for each $(i,j)$ of the $N\times N$ node pairs (see \cref{sec:apx_egt} for details). Triplet interaction operates only on the pairwise embeddings $\mathbf{e}_{ij}$.

\begin{figure}[t]
    \centering
    \includegraphics[width=0.75\linewidth]{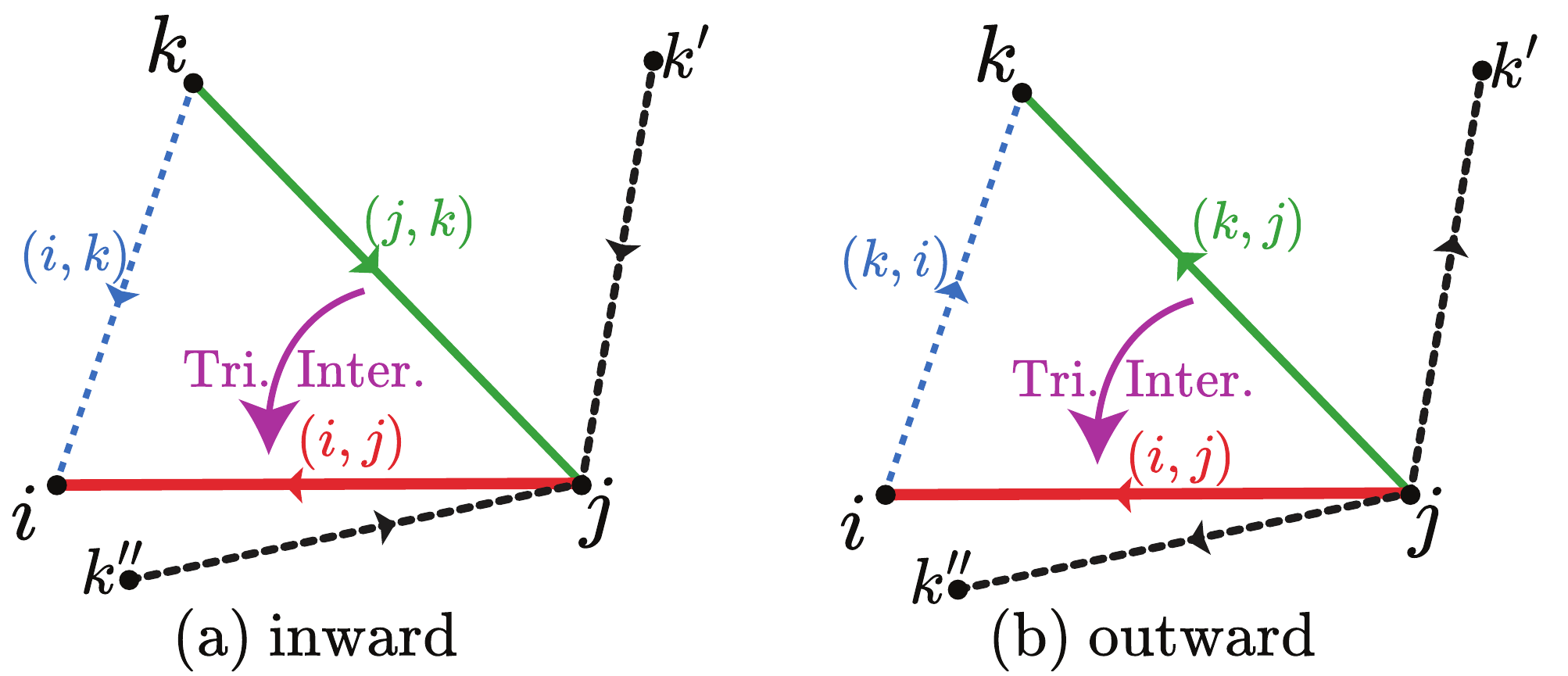}
    \caption{Triplet interaction allows direct communication between two adjacent pairs  $(i, j)$ and $(j, k)$, alleviating the bottleneck at the junction node $j$ and also takes into account the third pair $(i, k)$ within the 3-tuple $(i,j,k)$. (a) inward update (b) outward update.}
    \label{fig:triplet}
\end{figure}

TGT addresses the following important limitation of EGT -- it updates the pairwise embeddings $\mathbf{e}_{ij}$ solely based on the node embeddings $\mathbf{h}_i$ and $\mathbf{h}_j$. While this choice ensures a computational complexity of $O(N^2)$ like the original transformer, it constrains the model's expressivity to that of a 1-GWL test \citep{joshi2023expressive}. To improve upon this, we must move beyond 2nd-order interactions within the 2-tuple of nodes $(i,j)$ and consider 3rd-order interactions involving the 3-tuple of nodes $(i,j,k)$. Triplet interaction allows direct information flow to the pair $(i,j)$ from the neighboring pair $(j,k)$. To complete the 3-tuple, it also considers the pair $(i,k)$. As illustrated in \cref{fig:triplet}, the linear arrows between a pair of nodes denote the information flow in the node channels, whereas the curved arrow represents the information flow between pairwise embeddings due to triplet interaction. This interaction allows $(i,j)$ to aggregate all neighboring pairs $(j,k), (j,k'), (j,k''),\ldots$ without involving the junction node $j$, resolving the bottleneck at $j$. This 3rd-order interaction elevates the model's expressivity beyond 1-GWL, approaching that of 2-GWL, allowing it to model complex geometric relationships in graphs. But this comes at the cost of increased computational complexity. We only consider 3rd-order interactions which limits the complexity to $O(N^3)$. This is a good engineering choice since, as shown by \citet{li2024distance}, 3rd-order interactions are crucial for geometric understanding, yet $\ge$4th-order interactions add little/no benefit at much higher computational cost. Also, sub-cubic complexity is achievable with some concessions, as we will see in the next section. Some previous works have also used 3rd-order interactions via axial attention \citep{liu2022gem} or triangular update  \citep{lu2023highly,jumper2021highly} (see \cref{sec:apx_pair_to_pair} for details).

\begin{figure}[t]
    \centering
    \includegraphics[width=\linewidth]{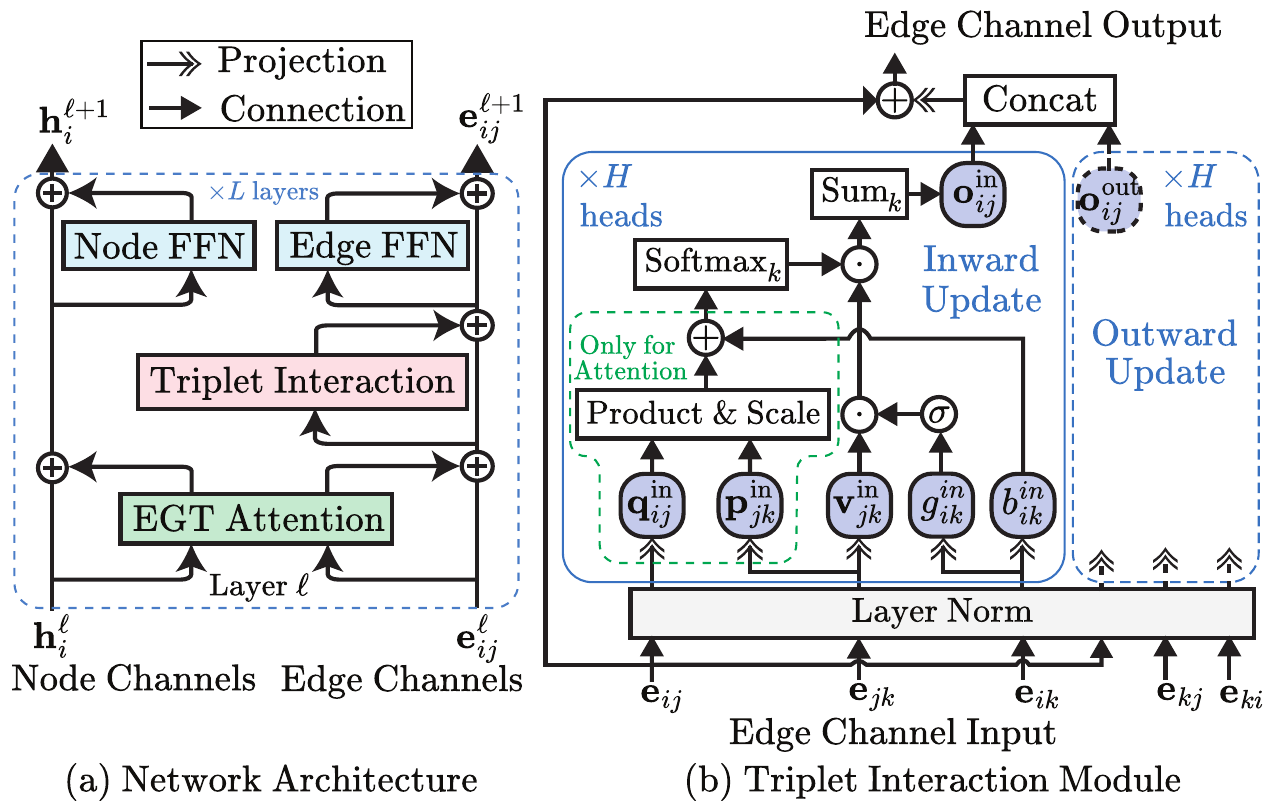}
    \vspace{-0.25in}
    \caption{(a) Triplet Graph Transformer (TGT) (b) Triplet Interaction Module (inward update shown in detail, outward is similar).}
    \label{fig:network}
\end{figure}

We introduce a novel triplet interaction module to the edge channels, between the pairwise attention block and the edge Feed Forward Network (FFN) block. This module follows the same pre-norm layer normalization and residual connection pattern as the rest of the network. The resultant Triplet Graph Transformer (TGT) architecture is shown in \cref{fig:network}(a). The triplet interaction module is shown in \cref{fig:network}(b). We propose two forms of interaction mechanisms called triplet attention and triplet aggregation and refer to the resultant variants as TGT-At and TGT-Ag, respectively.

\ssubsec{Triplet Attention (TGT-At)}
For a pair $(i, j)$, the triplet attention is computed as follows
\begin{align}
   \mathbf{o}_{ij}^{\mathrm{in}} &= \sum_{k=1}^N a_{ijk}^{\mathrm{in}} \mathbf{v}_{jk}^
   {\mathrm{in}} \label{eqn:tgt_at_in1}\\
   a_{ijk}^{\mathrm{in}} &= \softmax_k(\frac{1}{\sqrt{d}}\mathbf{q}_{ij}^{\mathrm{in}} \cdot \mathbf{p}_{jk}^{\mathrm{in}} + b_{ik}^{\mathrm{in}}) \times \sigma(g_{ik}^{\mathrm{in}}) \label{eqn:tgt_at_in2}
\end{align}
where the value vector $\mathbf{v}_{jk}^{\mathrm{in}}$ is derived from a learnable projection of the pairwise embedding $\mathbf{e}_{jk}$ and $a_{ijk}$ is the attention weight assigned to the pair $(j,k)$ by the pair $(i,j)$. This is done for multiple attention heads and $\mathbf{o}_{ij}^{\mathrm{in}}$ is the output of an attention head. $\mathbf{q}_{ij}^{\mathrm{in}}$ and $\mathbf{p}_{jk}^{\mathrm{in}}$ are the query and the key vectors derived from the pairwise embeddings $\mathbf{e}_{ij}$ and $\mathbf{e}_{jk}$, respectively. $b_{ik}^{\mathrm{in}}, g_{ik}^{\mathrm{in}}$ are scalars derived from the pairwise embeddings $\mathbf{e}_{ik}$ belonging to the third pair $(i,k)$ within the 3-tuple $(i,j,k)$, which participates by providing these bias and gating terms, respectively. The gating term is not strictly necessary but improves the performance of the model. We call this an inward update. Another parallel update, called outward update, is done by changing the direction of the aggregated pairs as follows:
\begin{align}
   \mathbf{o}_{ij}^{\mathrm{out}} &= \sum_{k=1}^N a_{ikj}^{\mathrm{out}} \mathbf{v}_{kj}^{\mathrm{out}} \label{eqn:tgt_at_out1}\\
    a_{ikj}^{\mathrm{out}} &= \softmax_k(\frac{1}{\sqrt{d}}\mathbf{q}_{ij}^{\mathrm{out}} \cdot \mathbf{p}_{kj}^{\mathrm{out}} + b_{ki}^{\mathrm{out}}) \times \sigma(g_{ki}^{\mathrm{out}}) \label{eqn:tgt_at_out2}
\end{align}
Finally, $\mathbf{o}_{ij}^{\mathrm{in}}$ and $\mathbf{o}_{ij}^{\mathrm{out}}$ for all heads are concatenated and the the pairwise embedding $\mathbf{e}_{ij}$ is updated from a learnable projection of the resultant. Triplet attention combines the strengths of axial attention \citep{liu2022gem} and triangular update \citep{jumper2021highly} in a single update. It is thus the most expressive form of interaction and outperforms both of the aforementioned mechanisms in our experiments. However, it has an $O(N^3)$ compute complexity.

\ssubsec{Triplet Aggregation (TGT-Ag)}
Triplet aggregation can be expressed as follows for the inward update:
\begin{align}
   \mathbf{o}_{ij}^{\mathrm{in}} &= \sum_{k=1}^N a_{ik}^{\mathrm{in}} \mathbf{v}_{jk}^{\mathrm{in}} \label{eqn:tgt_ag_in1}\\
   a_{ik}^{\mathrm{in}} &=  \softmax_k(b_{ik}^{\mathrm{in}})\times \sigma(g_{ik}^{\mathrm{in}}) \label{eqn:tgt_ag_in2}
\end{align}
Notice that it is a tensor multiplication between the weight matrix and the value matrix, each of which has only $O(N^2)$ elements and thus, has a subcubic complexity of $\approx O(N^{2.37})$ \citep{ambainis2015fast} which is much better than the cubic complexity of triplet attention. As a compromise, we have to remove the dependence of the weights on the junction node $j$. The weights are only determined by the pair $(i,k)$, due to removing the dot product term from the weights. Thus, this process is not an attention mechanism. We also have an outward update, and the final update is done by concatenating the inward and outward updates. Note that the weights are bounded and normalized (ignoring the gating term) unlike the triangular update in UniMol+ \citep{lu2023highly}. Also, the values being aggregated are vectors instead of scalars, which makes the process more expressive and efficient as only a few heads are required instead of many.

A comparison of different 3rd-order interaction mechanisms (previous methods and our triplet interactions) is shown in \cref{tab:triplet}. Triplet \emph{attention} is the most expressive because all other interactions are ablated versions of it, i.e., they can be derived by removing some terms from triplet attention (see \cref{eqn:tgt_at_in1,eqn:tgt_at_in2,eqn:tgt_at_out1,eqn:tgt_at_out2,eqn:tgt_ag_in1,eqn:tgt_ag_in2,eqn:axial_at1,eqn:axial_at2,eqn:tri_up1}). Triplet \emph{aggregation} is more expressive than triangular update because it aggregates vector values rather than scalar values. In our experiments, it outperforms triangular update while being more efficient. Axial attention is the least expressive because it does not include the pair $(i,k)$, and thus interaction within the 3-tuple is incomplete. In summary, triplet attention is the best-performing yet heavyweight method while triplet aggregation is more efficient and scalable at the cost of some performance. Both are better than previous methods.
\begin{table}[t]
    \centering
    \caption{Comparison of different 3rd-order interactions.}
    \scalebox{0.77}{
    \input{tables/triplet_comparison_colored.tex}
    }
    \newline
    \scriptsize{\tsup{1}\citet{liu2022gem}, \tsup{2}\citet{lu2023highly} }
    \label{tab:triplet}
\end{table}
\begin{figure}[t]
    \centering
    \includegraphics[width=\linewidth]{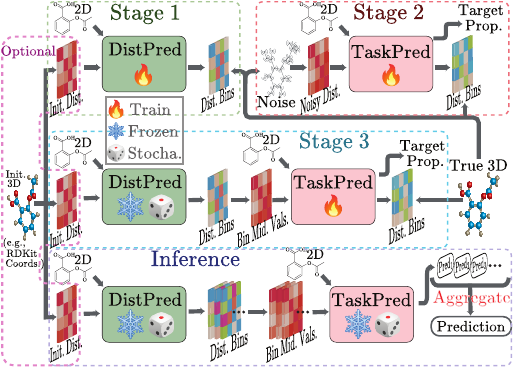}
    \vspace{-0.25in}
    \caption{The three stages of training and the stochastic inference.}
    \label{fig:train_infer}
\end{figure}

\ssubsec{Regularization Methods}
We introduce a new dropout \citep{srivastava2014dropout} method for triplet interaction, called triplet dropout. Following attention dropout \citep{zehui2019dropattention}, we randomly zero out the weights ($a_{ijk}$) of the triplet interaction mechanism by sampling a binary mask $M_{ijk}$ from a Bernoulli distribution with probability $p$ and multiplying it with the weights $a_{ijk}$. This is done for each attention head and both the inward and outward updates.

We also propose a new dropout method for the pairwise attention mechanism called source dropout. Instead of applying traditional attention dropout which randomly drops individual members of the attention matrix, we drop entire columns, i.e. key-value pairs, for all queries in all heads in a layer. This essentially makes some of the nodes ``invisible'' as information sources for the other nodes during this information exchange process. This is a stronger form of regularization than the traditional attention dropout, inspired by the structured dropout pattern proposed by \citet{hussain2023information}. It helps the model be robust to node degree variations in the input graphs, and thus more effective than traditional attention dropout for graphs.

\subsection{Training and Inference for Molecular Graphs}

Our training method for molecular property prediction consists of three stages. The first stage involves training the distance predictor which predicts interatomic distances from 2D molecular graphs. The second stage involves pretraining the task predictor which predicts molecular properties on noisy 3D structures, and the third stage involves finetuning it on the predicted distances. The three stages are shown in \cref{fig:train_infer} along with the inference process.

\ssubsec{Stage 1: Training the Distance Predictor}
We train a TGT distance predictor to predict interatomic distances for all atom pairs in the molecule. The model takes the 2D molecular graph as input and outputs a binned, clipped distance matrix. We directly use the output edge embeddings to predict pairwise distances. We found predicting distances up to 8\r{A} is enough for molecules. Cross-entropy loss is used to train the model. Optionally, an initial distance estimate, e.g., from RDKit coordinates can be used to improve accuracy.

Our reasoning for predicting distances instead of coordinates is that they can be directly used by the task predictor. We predict and utilize the full distance matrix which fully defines the geometry of the molecule up to a reflection (i.e., it cannot capture chirality). Thus, defining all pairwise distances also defines all the angles and dihedrals. Distances are also invariant to translation/rotation and have a small, easily learned value range. We found the accuracy of predicting individual distances is more important than strictly conforming distances to a 3D structure. Unlike geometric GNNs like GemNet \cite{gasteiger2021gemnet}, 3rd-order features like angles are not necessary, due to our novel triplet interaction (as shown by \citep{li2024distance}). Additionally, unlike UniMol+ \citep{lu2023highly}, no initial 3D structure is necessary for our approach, which is a significant advantage since the whole inference pipeline can be pure neural network-based and highly scalable. We predict binned distances instead of continuous values since distance distributions are often multimodal and skewed, and thus only fully captured by the model with a cross-entropy loss. This allows us to predict the most probable distance, i.e. the mode, which is more accurate than the mean or median predictions. Quantization noise due to binning does not affect the downstream task predictor which is robust to noise.

\ssubsec{Stage 2: Pretraining the Task Predictor}
\label{sec:pretrain}
We first pretrain a TGT task predictor on the noisy ground truth 3D structures (when available) rather than directly training on predicted distances. It makes the task predictor robust to noise in input distances and makes it adaptable to less accurate predicted distances. Similar to previous works \citep{godwin2021simple}, this also serves as an effective regularizer when we include a pairwise distance prediction head in the task predictor with a secondary objective of predicting ground truth binned distances by encouraging the edge channels to denoise the 3D structure. Without it, model accuracy stagnates or deteriorates by failing to provide useful supervision to these channels. The distance prediction secondary objective combines with the primary task prediction objective in a multitask learning setup and serves as a powerful regularization method and can even be incorporated when training directly on less accurate 3D data like RDKit coordinates.

We propose a novel input 3D noising method where instead of adding random Gaussian or uniform noise that disproportionately affects local versus global structure, we inject locally smoothed noise that better reflects distance noise characteristics. Specifically, atoms in closer proximity move together, while far-apart atoms move more independently. This can be expressed as:
\begin{align}
    \mathbf{r}_i' = \mathbf{r}_i + \sum_{j=1}^N e^{-\frac{\lVert\mathbf{r}_i - \mathbf{r}_j\rVert}{\nu}} \mathbf{u}_j
    \text{ ;  where } \mathbf{u}_j \sim \mathcal{N}(0, \sigma^2 \mathbf{I})
\end{align}
Here, $\mathbf{r}_i$ is the ground truth 3D coordinate of atom $i$, $\mathbf{r}_i'$ is the noised coordinate, and $\mathbf{u}_j$ is the 3D noise vector corresponding to atom $j$. The nature of the noise can be controlled by tuning the smoothing parameter $\nu$ and the noise variance $\sigma^2$. We found that $\nu=1$\r{A} is a good choice for most cases, while $\sigma$ can be tuned to set the noise level. 

\ssubsec{Stage 3: Finetuning the Task Predictor on Predicted Distances}
Before inference, the task predictor must adjust to using predicted interatomic distances from the frozen distance predictor. During this finetuning process, we keep the distance predictor in \emph{stochastic} mode with active dropout during inference. Although we choose the highest probability distance bin, this enables sampling multiple predictions for the same input, like a probabilistic model, and serves as effective data augmentation, regularizing the finetuning process. During finetuning, we maintain the distance prediction objective from pretraining, although optionally with reduced weight which continues to encourage noise robustness.

\subsection{Stochastic Inference}
During inference, we use stochastic distance predictions and also leverage the task predictor in stochastic mode (i.e., dropouts are active) to predict target tasks. The task predictor makes predictions on each distance sample, which are aggregated via mean, median, or mode. It allows the task predictor to process different structural variations to produce a robust final prediction. This is reminiscent of using multiple conformations to account for structural flexibility. This process is non-iterative, embarrassingly parallel, and highly scalable across multiple GPUs. Only $\approx10$ samples produce very good results, which further improves monotonically with more samples. The prediction distribution also enables uncertainty estimation which is especially useful to guide the search for new molecules with desired properties.

\begin{table}[t]
    \centering
    \caption{Results on PCQM4Mv2.}
    \scalebox{0.77}{
        \input{tables/pcqm_results.tex}
    }
    \newline
    \scriptsize{\tsup{1}\tciteGINE{}, \tsup{2}\tciteGCN{}, \tsup{3}\tciteGIN{}, \tsup{4}\tciteDeeperGCN{}, \tsup{5}\tciteTokenGT{}, \tsup{6}\tciteGRPE{}, \tsup{7}\tciteGraphormer{}, \tsup{8}\tciteGPS{}, \tsup{9}\tciteEGT{}, \tsup{10}\tciteGEMtwo{}, \tsup{11}\tciteTransfoM{}, \tsup{12}\tciteGPSPP{}, \tsup{13}\tciteUniMolP{}, \tsup{14}\tciteVN{} }
    \label{tab:pcqm}
    \vspace{-0.1in}
\end{table}

\begin{table}[t]
    \centering
    \caption{Average results on the OC20 IS2RE task.}
    \scalebox{0.77}{
\input{tables/oc20_avg_results.tex}

    }
    \newline
    \scriptsize{\tsup{1}\tciteSchnet{}, \tsup{2}\tciteDimeNetPP{}, \tsup{3}\tciteSphereNet{}, \tsup{4}\tciteGNS{}, \tsup{5}\tciteGraphormerthreeD{}, \tsup{6}\tciteEquiformer{}, \tsup{7}\tciteDRFormer{}, \tsup{8}\tciteMoleformer{}, \tsup{9}\tciteUniMolP{}, \tsup{10}\tciteNN {} }
    \label{tab:oc20_avg}
\end{table}

\begin{table}[t]
    \centering
    \caption{A breakdown of performance of top two models -- UniMol+ \citeUniMolP{} and our TGT-At on different splits of the OC20 IS2RE validation and test datasets.}
    \scalebox{0.77}{
\input{tables/oc20_best2_results_flat.tex}
    }
    \label{tab:oc20_best2}
\end{table}

\section{Experiments}
Our experiments are designed to validate several key aspects of our proposed model and training approach. Firstly, we demonstrate the performance and scalability of our approach on large quantum chemistry datasets PCQM4Mv2 \citeOGBLSC{} and OC20 \citeOCtwenty{}. Next, we evaluate the transfer learning capabilities of our models, by finetuning our task predictor from PCQM4Mv2 to related quantum chemistry tasks on the QM9 \citeQMnine{} dataset. We also transfer our distance predictor from PCQM4Mv2 to molecular property prediction on MOLPCBA \citeOGB{} and drug discovery dataset LIT-PCBA \citeLITPCBA{}. Finally, we showcase the utility of our triplet interaction mechanisms for graph learning in general by evaluating it on the traveling salesman problem task on the TSP dataset by \tciteBenchmarkingG{}. The PyTorch \citePyTorch{} library was used to implement the models. The training was done with mixed-precision computation on 4 nodes, each with 8 NVIDIA Tesla V100 GPUs (32GB RAM/GPU), and two 20-core 2.5GHz Intel Xeon CPUs (768GB RAM). The hyperparameters, training, and dataset details are provided in \cref{sec:apx_more_training}. Our code is available at \url{https://github.com/shamim-hussain/tgt}.

\subsection{Large-scale Quantum Chemical Prediction}
\ssubsec{PCQM4Mv2}
PCQM4Mv2 comprising 4 million molecules, is a part of the OGB-LSC datasets \citeOGBLSC{}. The primary objective is predicting the HOMO-LUMO gap. The performance of the distance predictor is tuned on a random subset of 5\% of the training data which we call validation-3D. The training of our TGT-At model takes $\approx 32$ A100 GPU-days, slightly less than the training time of UniMol+ \citeUniMolP{}, which takes 40 A100 GPU-days.

The results of our experiments are presented in \cref{tab:pcqm} in terms of Mean Absolute Error (MAE) in meV unit. Our TGT-At model achieves the best performance on the PCQM4Mv2 dataset, outperforming the previous SOTA UniMol+ model by a significant margin of 2.2 meV. It is worth highlighting that  UniMol+ uses RDKit coordinates as input which is optional for our model. We see that even without RDKit coordinates, i.e., with a pure neural approach, our model outperforms UniMol+ by a fair margin. Hence, we currently hold the top positions on the PCQM4Mv2 leaderboard for both RDKit-aided and pure neural approaches. The TGT-Ag model also exhibits good performance, securing the second-best position after TGT-At. TGT-Agx2 reduces parameter count by half by sharing parameters in consecutive layers yet still outperforms UniMol+. Our performance gains over other models stem from two factors -- not only a superior architecture but also better training and inference. This is evidenced by the success of a basic EGT 2-stage model under our training and inference paradigm.

\ssubsec{Open Catalyst 2020 IS2RE}
The Open Catalyst 2020 Challenge \citeOCtwenty{} is aimed at predicting the adsorption energy of molecules on catalyst surfaces using machine learning. We focus on the IS2RE (Initial Structure to Relaxed Energy) task, where the model is provided with an initial DFT structure of the crystal and adsorbate, which interact with each other to reach the relaxed structure when the relaxed energy of the system is measured. We exclusively use the IS2RE dataset and limit the number of atoms to a maximum of 64 by cropping/sampling based on distances to the adsorbate atoms. It takes $\approx 32$ A100 GPU-days to train the model, which is significantly lower than the 112 GPU-days used by UniMol+, due to using much smaller sized graphs and also our more efficient training method.

The results for the IS2RE task are shown in \cref{tab:oc20_avg} in terms of MAE (in meV) and percent Energy within a Threshold (EwT) of 20 meV. We see that TGT-At performs better than the SOTA UniMol+ model while using significantly less compute, both for training and inference. TGT-Ag performs second best and still outperforms other direct methods while being significantly faster. The IS2RE evaluation is carried out over multiple sub-splits of the validation and test datasets - ID (In Domain) and OOD (Out of Domain) Adsorbates, Catalyst, or Both. The breakdown for the best two models -- UniMol+ and TGT-At are presented in \cref{tab:oc20_best2}. We see that TGT-At outperforms UniMol+ on OOD splits which are of more importance, and overall performs slightly better when both MAE and EwT are considered. Thus our TGT-At model secures the spot of the best-performing direct method on the OC20 IS2RE leaderboard.

\begin{figure}[t]
    \centering
    \includegraphics[width=\linewidth]{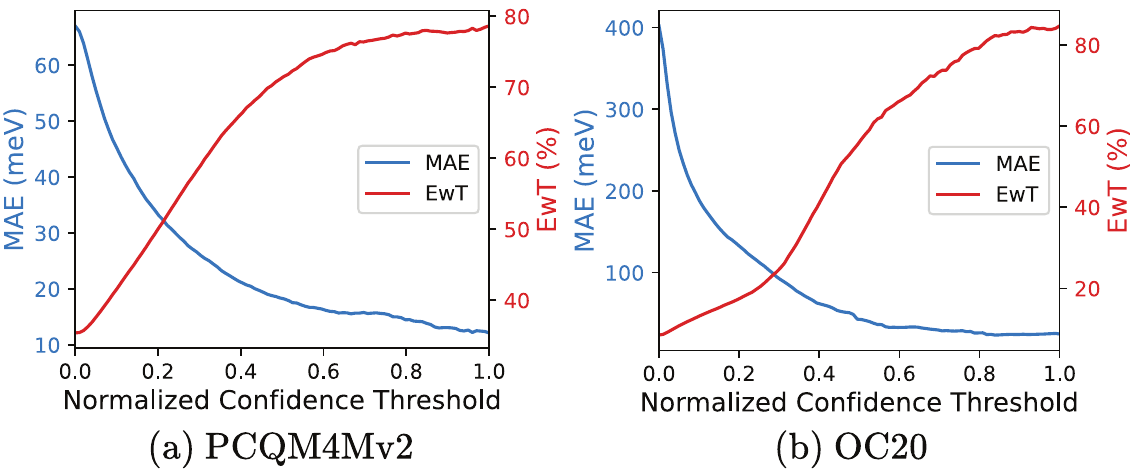}
    \vspace{-0.3in}
    \caption{Normalized Confidence Threshold vs MAE (meV) and EwT (\%) on (a) PCQM4Mv2 and (b) OC20 IS2RE validation sets.}
    \label{fig:uncertainty}
    \vspace{-0.1in}
\end{figure}

\ssubsec{Uncertainty Estimation}
Our stochastic inference method allows us to draw multiple sample predictions for each input, which can be used to estimate the uncertainty of our predictions by looking at the spread of the samples. Specifically, we use the reciprocal of standard deviation as a confidence measure, as shown in \cref{fig:uncertainty}. For better visualization, we normalize the confidence measure to the range $[0,1]$. We plot the MAE and EwT of validation graphs, filtered by confidence threshold. We can see that performance monotonically improves with higher confidence -- evidenced by lower MAE and higher EwT. This shows that the confidence of a prediction has a strong positive correlation with its accuracy. This can be useful for real-world applications like drug discovery and material design.

\subsection{Transfer Learning}
Our model learns two different forms of knowledge during the large-scale training on the PCQM4Mv2 dataset. The distance predictor learns to predict interatomic distances from 2D molecular graphs and the task predictor learns to predict the quantum chemical property of HOMO-LUMO gap from 3D molecular graphs. Thus we test the transfer of knowledge in two different settings in this section.

\ssubsec{Finetuning on QM9} To highlight the transfer of knowledge to related quantum chemical prediction tasks we finetune the task predictor from PCQM4Mv2 on the QM9 \citeQMnine{} dataset. The ground truth 3D coordinates are provided on this dataset which can be used during inference, so the distance predictor is not required. We report finetuning performance on a subset of 7 tasks from the 12 tasks in QM9 in \cref{tab:qm9}. We get comparable results with TGT-Ag and TGT-At, so we only report results for TGT-Ag to save compute. We see that TGT-Ag archives SOTA results and outperforms other models by a significant margin in predicting the HOMO ($\epsilon _{H}$), LUMO($\epsilon _{L}$), and the HOMO-LUMO gap ($\Delta \epsilon$). This is because these tasks are directly related to the pretraining task. We also achieve SOTA results for $\alpha$ and $C_v$ and perform satisfactorily on the other two tasks -- demonstrating a positive transfer of knowledge to these tasks. Notably, we outperform Transformer-M \citeTransfoM{}, another transformer model pretrained on PCQM4Mv2 due to our novel triplet interaction mechanism.

\begin{table}
    \centering
    \caption{Results (MAE\lob{}) on the QM9 dataset.}
    \scalebox{0.75}{
        \input{tables/qm9_results.tex}
    }
    \newline
    \scriptsize{\tsup{1}\tciteGraphMVP{}, \tsup{2}\tciteGEM{}, \tsup{3}\tciteInfoM{}, \tsup{4}\tciteMGP{}, \tsup{5}\tcitePhysNet{}, \tsup{6}\tciteSchnet{}, \tsup{7}\tciteCormorant{}, \tsup{8}\tciteDimeNetPP{}, \tsup{9}\tcitePaiNN{}, \tsup{10}\tciteEGNN{}, \tsup{11}\tciteNN{}, \tsup{12}\tciteSphereNet{}, \tsup{13}\tciteSEGNN{}, \tsup{14}\tciteEQGAT{}, \tsup{15}\tciteSEthreeT{}, \tsup{16}\tciteTorchMD{}, \tsup{17}\tciteEquiformer{}, \tsup{18}\tciteTransfoM{}}
    \label{tab:qm9}
    \vspace{-0.1in}
\end{table}

\begin{table}[t]
    \centering
    \caption{Results on MOLPCBA.}
    \scalebox{0.77}{
\input{tables/molpcba_results.tex}
    }
    \newline
    \scriptsize{\tsup{1}\tciteDeeperGCN{}, \tsup{2}\tciteNPA {}, \tsup{3}\tciteDGN{}, \tsup{4}\tciteGINE{}, \tsup{5}\tcitePHC{}, \tsup{6}\tciteGIN{}, \tsup{7}\tciteGraphormer{}, \tsup{8}\tciteEGT{}, \tsup{9}\tciteVN{}, \tsup{10}\tciteFLAG{}}
    \label{tab:molpcba}
    \vspace{-0.1in}
\end{table}

\begin{table}[t]
    \centering
    \caption{Average results on LIT-PCBA.}
    \scalebox{0.77}{
\input{tables/litpcba_avg_results_trunc.tex}
    }
    \newline
    \scriptsize{\tsup{1}\tciteGCN{}, \tsup{2}\tciteGAT{}, \tsup{3}\tciteFPGNN{}, \tsup{4}\tciteEGT{}, \tsup{5}\tciteGEM{}, \tsup{6}\tciteGEMtwo{}}
    \label{tab:litpcba}
    \vspace{-0.05in}
\end{table}

\begin{table}[t]
    \centering
    \caption{Results on TSP.}
    \scalebox{0.77}{
\input{tables/tsp_results.tex}
    }
    \newline
    \scriptsize{\tsup{1}\tciteGCN{}, \tsup{2}\tciteGIN{}, \tsup{3}\tciteGAT{}, \tsup{4}\tciteGatedGCN{}, \tsup{5}\tciteGraphormer{}, \tsup{6}\tciteARGNP{}, \tsup{7}\tciteEGT{}}
    \label{tab:tsp}
\end{table}

\ssubsec{Molecular Property Prediction}
Since 3D geometric information is valuable for molecular property prediction, we use our pretrained distance predictor (without RDKit) to provide estimations of interatomic distances to models on the MOLPCBA \citeOGB{} molecular property prediction and the LIT-PCBA \citeLITPCBA{} drug discovery benchmarks. These datasets do not have ground truth 3D information. So, we do not finetune the distance predictor on them but rather use it as a frozen feature extractor. The task predictor is trained from scratch and takes the predicted distances as input.

The results for MOLPCBA are presented in \cref{tab:molpcba} in terms of test Average Precision (\%) which evaluates the performance of the model in a multitask setting of predicting 128 different binary molecular properties. We see that both EGT and TGT-Ag models trained from scratch with RDKit coordinates get good results but if we use our pretrained TGT-AT-DP (``-DP'' stands for distance predictor), we get the best results. We also see that our model outperforms the SOTA pretrained Graphormer model by a significant margin.

On the LIT-PCBA dataset, we report on an average ROC-AUC (\%) on 7 separate tasks of predicting interactions with proteins in \cref{tab:litpcba} (a breakdown is provided in \cref{tab:apx_litpcba_val} in the appendix). We see that EGT with TGT-At-DP matches the SOTA pretrained GEM-2. Both of these experiments demonstrate that our pretrained TGT-At-DP can provide valuable 3D information to the task predictor, even though it is trained on a different dataset. We also see that our TGT-At-DP which is trained on DFT coordinates can provide more useful 3D information than RDKit coordinates.

\subsection{Traveling Salesman Problem}
We also show the utility of our proposed triplet interaction mechanism beyond molecular graphs and for general-purpose graph learning by targetting the Traveling Salesman Problem benchmark dataset by \tciteBenchmarkingG{} which consists of 12000 K-NN graphs of 50-500 2D points in the unit square. The task is to predict which of the edges of the K-NN graph are part of the optimal tour. Since distance prediction is not required, we train a single-stage model that performs binary edge classification, and no pretraining/fine-tuning is involved. The model receives both coordinates (node features) and pairwise distances (edge features) as input. According to the specification of this benchmarking dataset, the task must be performed with a given parameter budget of either 100K or 500K.

The results are presented in \cref{tab:tsp} in terms of test F1 score (\%) where We get a significant improvement when we use our TGT-Ag and TGT-At models. This shows that our triplet interaction mechanism is very useful for solving the TSP task. We do not evaluate the TGT-At model for the larger 500K parameter budget due to memory constraints. We show the performance of the TGT-Ag model can be further improved by using repeated layers with shared parameters, dubbed as TGT-Agx2, TGT-Agx3, and TGT-Agx4. This shows the effectiveness of our triplet interaction module in an iterative setting.

\subsection{Ablation Study}
In \cref{tab:triplet_abl}, we compare our proposed triplet interaction methods with the previously proposed 3rd-order mechanisms -- axial attention and triangular update, which can be thought of as ablated variants of our method. We also compare with ungated variants of our methods (i.e., without gating terms). We compare the cross-entropy losses of the distance predictors (a good indicator of the downstream performance) with different 3rd-order interaction mechanisms on the PCQM4Mv2 validation-3D set. We normalize training time with respect to the no 3rd-order interaction scenario. We see that triplet attention performs best but is expensive. Triplet aggregation performs better than both axial attention and triangular update, yet is more efficient. The ungated variants perform slightly worse but are also slightly more efficient. We also see that the model with no 3rd-order interaction performs the worst by a significant margin, which shows its importance for distance prediction.

\begin{table}[t]
    \centering
    \caption{Distance prediction performance of different 3rd-order interaction mechanisms and training times on PCQM4Mv2.}
    \scalebox{0.77}{
        \input{tables/triplet_dp_compare.tex}
    }
    \label{tab:triplet_abl}
\end{table}
\begin{table}[t]
    \centering
    \caption{Ablation Study on PCQM4Mv2.}
    \scalebox{0.77}{
        \input{tables/ablation.tex}
    }
    \label{tab:ablation}
\end{table}

\cref{tab:ablation} shows a detailed ablation study to test the effectiveness of our triplet interaction mechanism and other proposed optimizations within our model and training framework. Results are shown for a smaller 12-layer model on PCQM4Mv2. We see that to take full advantage of the input 3D information (e.g., RDKit coordinates), we also need the denoising distance prediction objective. Local smoothing of the input noise improves this process. Source dropout proves to be a better alternative to attention dropout. Incorporating DFT distance predictor, and pretraining on noisy DFT coordinates both lead to significant improvements individually and even greater improvements when combined. Finally, a significant leap comes from triplet attention.

\section{Limitations}
The main limitations of TGT lie in the computational complexity ($\geq O(N^{2.37})$) of the triplet interaction mechanism which is higher than the  $O(N^2)$ complexity of base transformers like EGT. This is because, while EGT only considers pairwise interactions, TGT considers 3rd order interactions. However, we predict that this disadvantage can be alleviated by incorporating sparse and/or low-rank interactions. In this work, we focus on maximizing the performance of the model, and any exploration of the trade-off between performance and complexity is left for future work.

\section{Conclusion}
In this work, we introduce the Triplet Graph Transformer (TGT) architecture, which incorporates the 3rd-order triplet interaction mechanism. This significantly improves the modeling of geometric dependencies in graph transformers. We proposed two forms of triplet interactions -- an attention-based mechanism with maximum expressivity, and an aggregation-based mechanism with greater efficiency and scalability. Additionally, we put forth a two-stage framework involving separate distance predictor and property predictor models for molecular graphs. Our distance predictor directly predicts interatomic distances from 2D graphs during inference, eliminating the need for property prediction on 2D information only. The three-stage training methodology with a stochastic inference scheme enables fast and accurate predictions, significantly advancing over previous iterative refinement approaches, and allows for uncertainty quantification in the prediction. Through extensive experiments, we demonstrate state-of-the-art predictive accuracy on quantum chemical datasets and the transferability of our distance predictor to molecular property prediction. Moreover, the superior performance of TGT on the TSP task indicates the broad applicability of our proposed triplet interactions. In future work, we plan to explore the use of our triplet interaction mechanism for other graph learning tasks. We also plan to evaluate its performance in other molecular graph learning tasks like molecule and conformation generation. Additionally, we aim to further investigate improving the compute and memory efficiency of triple interaction.

\section*{Acknowledgements}
This work was supported by the Rensselaer-IBM AI Research Collaboration, part of the IBM AI Horizons Network.

\section*{Impact Statement}
Our work aims to improve chemical property prediction, which can accelerate the discovery of new beneficial materials and drugs. This has the potential to greatly benefit society by enabling the development of cheaper, safer, and more effective medicines as well as advanced materials for clean energy and other applications. Our model can be used to efficiently solve the Traveling Salesman Problem for improved routing and logistics planning in transportation. However, though increased automation can reduce costs, over-reliance on AI systems without human oversight raises concerns about accountability and bias. We have made the code open-source so that domain experts can further improve and validate the models before real-world deployment.

\bibliography{citations/citations, citations/dnn, citations/new}
\bibliographystyle{icml2024}
\balance

\newpage
\appendix
\onecolumn

\input{appendix.tex}


\end{document}

%% file: citations.tex
\newcommand{\citeEdgeVRP}{\citep{liu2023edge}}

\newcommand{\citeGraphormer}{\citep{ying2021transformers}}

\newcommand{\citeEGT}{\citep{hussain2022global}}

\newcommand{\citeGPS}{\citep{rampavsek2022recipe}}
\newcommand{\citeGPSPP}{\citep{masters2022gps++}}
\newcommand{\citeGRPE}{\citep{park2022grpe}}

\newcommand{\citeGraphormerthreeD}{\citep{shi2022benchmarking}}
\newcommand{\citeUniMolP}{\citep{lu2023highly}}

\newcommand{\citeGEMtwo}{\citep{liu2022gem}}

\newcommand{\citeTransfoM}{\citep{luo2022one}}
\newcommand{\citeSAN}{\citep{kreuzer2021rethinking}}

\newcommand{\citeSVM}{\citep{cortes1995support}}
\newcommand{\citeRndForest}{\citep{liaw2002classification}}
\newcommand{\citeXGBoost}{\citep{chen2016xgboost}}
\newcommand{\citeNaiveBayes}{\citep{duda1973pattern}}

\newcommand{\citeOGB}{\citep{hu2020open}}
\newcommand{\citeOGBLSC}{\citep{hu2021ogb}}
\newcommand{\citeOCtwenty}{\citep{chanussot2021open}}
\newcommand{\citeLITPCBA}{\citep{tran2020lit}}
\newcommand{\citeQMnine}{\citep{ramakrishnan2014quantum}}

\newcommand{\citeRDKit}{\citep{landrum2013rdkit}}
\newcommand{\citePyTorch}{\citep{paszke2019pytorch}}
\newcommand{\citeMMFF}{\citep{halgren1996merck}}

\newcommand{\tciteBenchmarkingG}{\citet{dwivedi2020benchmarking}}
\newcommand{\tciteGEMtwo}{\citet{liu2022gem}}

\newcommand{\tciteGCN}{\citet{kipf2016semi}}

\newcommand{\tciteGIN}{\citet{xu2018powerful}}
\newcommand{\tciteGINE}{\citet{brossard2020graph}}
\newcommand{\tciteGAT}{\citet{velivckovic2017graph}}

\newcommand{\tciteGatedGCN}{\citet{bresson2017residual}}
\newcommand{\tciteNPA}{\citet{corso2020principal}}
\newcommand{\tciteDGN}{\citet{beani2021directional}}
\newcommand{\tciteGraphormer}{\citet{ying2021transformers}}
\newcommand{\tciteARGNP}{\citet{cai2022automatic}}
\newcommand{\tciteEGT}{\citet{hussain2022global}}
\newcommand{\tciteDeeperGCN}{\citet{li2020deepergcn}}
\newcommand{\tcitePHC}{\citet{le2021parameterized}}
\newcommand{\tciteVN}{\citet{gilmer2017neural}}
\newcommand{\tciteFLAG}{\citet{kong2020flag}}
\newcommand{\tciteNN}{\citet{godwin2021simple}}
\newcommand{\tciteGPS}{\citet{rampavsek2022recipe}}
\newcommand{\tciteGPSPP}{\citet{masters2022gps++}}
\newcommand{\tciteGRPE}{\citet{park2022grpe}}
\newcommand{\tciteTokenGT}{\citet{kim2022pure}}

\newcommand{\tciteGNS}{\citet{kumar2022gns}}
\newcommand{\tciteGraphormerthreeD}{\citet{shi2022benchmarking}}
\newcommand{\tciteUniMolP}{\citet{lu2023highly}}
\newcommand{\tciteGraphMVP}{\citet{liu2021pre}}
\newcommand{\tciteGEM}{\citet{fang2021chemrl}}
\newcommand{\tciteInfoM}{\citet{stark20223d}}
\newcommand{\tciteMGP}{\citet{jiao20223d}}
\newcommand{\tciteSchnet}{\citet{schutt2017schnet}}
\newcommand{\tcitePhysNet}{\citet{unke2019physnet}}
\newcommand{\tciteCormorant}{\citet{anderson2019cormorant}}
\newcommand{\tciteSEthreeT}{\citet{fuchs2020se}}

\newcommand{\tciteDimeNetPP}{\citet{gasteiger2020dimenetpp}}
\newcommand{\tcitePaiNN}{\citet{schutt2021equivariant}}
\newcommand{\tciteTorchMD}{\citet{tholke2022torchmd}}
\newcommand{\tciteEGNN}{\citet{satorras2021n}}
\newcommand{\tciteSEGNN}{\citet{brandstetter2021geometric}}
\newcommand{\tciteEQGAT}{\citet{le2022equivariant}}
\newcommand{\tciteSphereNet}{\citet{liu2021spherical}}

\newcommand{\tciteMoleformer}{\citet{yuan2023molecular}}
\newcommand{\tciteDRFormer}{\citet{wang2023dr}}
\newcommand{\tciteEquiformer}{\citet{tholke2021equivariant}}
\newcommand{\tciteTransfoM}{\citet{luo2022one}}

\newcommand{\tciteFPGNN}{\citet{cai2022fp}}
\newcommand{\tciteSVM}{\citet{cortes1995support}}
\newcommand{\tciteRndForest}{\citet{liaw2002classification}}
\newcommand{\tciteXGBoost}{\citet{chen2016xgboost}}
\newcommand{\tciteNaiveBayes}{\citet{duda1973pattern}}

\newcommand{\tciteGemNet}{\citet{gasteiger2021gemnet}}

%% file: tables/triplet_comparison_colored.tex
\setlength{\tabcolsep}{4pt}
\begin{tabular}{l|cc|cc}
	\toprule
	             & \tcap{Axial Att.\tsup{1}} & \tcap{Tria. Update\tsup{2}}  & \tcap{Triplet Agg.}     & \tcap{Triplet Att.}     \\
	\midrule
	Normalized?  & \colgrn{\mtick}           & \colred{\mcross}             & \colgrn{\mtick}         & \colgrn{\mtick}         \\
	Gated?       & \colred{\mcross}          & \colgrn{\mtick}              & \colgrn{\mtick}         & \colgrn{\mtick}         \\
	Attention?   & \colgrn{\mtick}           & \colred{\mcross}             & \colred{\mcross}        & \colgrn{\mtick}         \\
	Values are   & \colgrn{Vectors}          & \colred{Scalars}             & \colgrn{Vectors}        & \colgrn{Vectors}        \\
	Participants & \colred{$ij, jk$}         & \colgrn{$ij, jk, ik$}        & \colgrn{$ij, jk, ik$}   & \colgrn{$ij, jk, ik$}   \\
	Weighted by  & \colred{$ij, jk$}         & \colred{$ik$}                & \colred{$ik$}           & \colgrn{$ij, jk, ik$}   \\
	Complexity   & \colred{$O(N^3)$}         & \colgrn{$O(N^{2.37})$}       & \colgrn{$O(N^{2.37})$}  & \colred{$O(N^3)$}       \\
	Expressivity & \colred{Worst}            & \colgrn{Good}                & \colgrn{Better}         & \colgrn{Best}           \\
	\bottomrule
\end{tabular}

%% file: tables/pcqm_results.tex
\setlength{\tabcolsep}{4pt}
\begin{tabular}{lccc}
	\toprule
	                              &                & \tcap{Val. MAE\lob{}}  & \tcap{Test-dev MAE\lob{}}     \\
	\tcap{Model}                  & \tcap{\#Param} & \tcap{(meV)}             & \tcap{(meV)}                \\\midrule
	GINE\tsup{1}-VN\tsup{14}      & 13.2M          & 116.7                    & -                           \\
	GCN\tsup{2}-VN\tsup{14}       & 4.9M           & 115.3                    & 115.2                       \\
	GIN\tsup{3}-VN\tsup{14}       & 6.7M           & 108.3                    & 108.4                       \\
	DeeperGCN\tsup{4}-VN\tsup{14} & 25.5M          & 102.1                    & -                           \\\midrule
	TokenGT\tsup{5}               & 48.5M          & 91.0                     & 91.9                        \\
	GRPE\tsup{6}                  & 118.3M         & 86.7                     & 87.6                        \\
	Graphormer\tsup{7}            & 47.1M          & 86.4                     & -                           \\
	GraphGPS\tsup{8}              & 13.8M          & 85.2                     & 86.2                        \\
	EGT\tsup{9}                   & 89.3M          & 85.7                     & 86.2                        \\
	GEM-2\tsup{10} (+RDKit)       & 32.1M          & 79.3                     & 80.6                        \\
	Transformer-M\tsup{11}        & 69M            & 77.2                     & 78.2                        \\
	GPS++\tsup{12}                & 44.3M          & 77.8                     & 72.0                        \\\midrule
	Uni-Mol+\tsup{13} (+RDKit)    & 77M            & 69.3                     & 70.5                        \\\midrule
	EGT\tsup{9} (2 Stage+RDKit)   & 189M           & 69.0                     & -                           \\\midrule
	TGT-Agx2 (+RDKit)             & 95M            & 68.2                     & -                           \\
	TGT-Ag (+RDKit)               & 192M           & \rgd{67.9}               & -                           \\
	TGT-At                        & 203M           & 68.6                     & \rgd{69.8}                  \\
	TGT-At (+RDKit)               & 203M           & \rbs{67.1}               & \rbs{68.3}                  \\
	\bottomrule
\end{tabular}

%% file: tables/oc20_avg_results.tex
\begin{tabular}{lcccc}
	\toprule
	                                & \multicolumn{2}{c}{\tcap{Val. Avg.}}                              & \multicolumn{2}{c}{\tcap{Test Avg.}}    \\ \cmidrule{2-5}
	                                & \tcap{MAE\lob{}}                & \tcap{EwT\hib{}}                & \tcap{MAE\lob{}} & \tcap{EwT\hib{}}     \\
	\tcap{Model}                    & \tcap{(meV)}                    & \tcap{(\%)}                     & \tcap{(meV)}       & \tcap{(\%)}        \\ \midrule
	SchNet\tsup{1}                  & 666.0                           & 2.65                            & 684.8              & 2.61               \\
	DimeNet++\tsup{2}               & 621.7                           & 3.42                            & 631.0              & 3.21               \\
	SphereNet\tsup{3}               & 602.4                           & 3.64                            & 618.8              & 3.32               \\
	GNS\tsup{4}+NN\tsup{10}         & 480.0                           & -                               & 472.8              & 6.51               \\ \midrule
	Graphormer-3D\tsup{5}           & 498.0                           & -                               & 472.2              & 6.10               \\
	EquiFormer\tsup{6}+NN\tsup{10}  & 441.0                           & 6.04                            & 466.0              & 5.66               \\
	DRFormer\tsup{7}                & 442.5                           & 6.84                            & 450.9              & 6.48               \\
	Moleformer\tsup{8}              & 460.0                           & 5.48                            & 458.5              & 6.48               \\
	Uni-Mol+\tsup{9}                & \rgd{408.8}                     & \rgd{8.61}                      & \rbs{414.3}        & \rgd{8.23}         \\\midrule
	TGT-Ag                          & 423.7                           & \rgd{8.64}                      & -                  & -                  \\
	TGT-At                          & \rbs{403.0}                     & \rbs{8.82}                      & \rgd{414.7}        & \rbs{8.30}         \\
	\bottomrule
\end{tabular}

%% file: tables/oc20_best2_results_flat.tex
\begin{tabular}{lcccc}
	\toprule
	                     & \multicolumn{2}{c}{\tcap{Val. Avg.}}        & \multicolumn{2}{c}{\tcap{Test Avg.}}\\ \cmidrule{2-5}
	Split                & \tcap{Uni-Mol+}  & \tcap{TGT-At}            & \tcap{Uni-Mol+} & \tcap{TGT-At}     \\ \midrule
	ID MAE\lob{}         & \rbs{379.5}      & 381.3                    & \rbs{374.5}     & 379.6             \\
	ID EwT\hib{}         & \rbs{11.15}      & \rbs{11.15}              & 11.29           & \rbs{11.50}       \\ \midrule
	OOD Ads. MAE\lob{}   & 452.6            & \rbs{445.4}              & 476.0           & \rbs{471.8}       \\
	OOD Ads. EwT\hib{}   & 6.71             & \rbs{6.87}               & \rbs{6.05}      & 5.70              \\ \midrule
	OOD Cat. MAE\lob{}   & 401.1            & \rbs{391.7}              & \rbs{398.0}     & 399.0             \\
	OOD Cat. EwT\hib{}   & 9.90             & \rbs{10.47}              & 9.53            & \rbs{9.84}        \\ \midrule
	OOD Both MAE\lob{}   & 402.1            & \rbs{393.6}              & 408.6           & \rbs{408.4}       \\
	OOD Both EwT\hib{}   & 6.68             & \rbs{6.80}               & 6.06            & \rbs{6.17}        \\
	\bottomrule
\end{tabular}

%% file: tables/qm9_results.tex
\begin{tabular}{lccccccc}
	\toprule
	\tcap{Method}          & $\mathbf{\mu}$       & $\mathbf{\alpha}$    & $\mathbf{\epsilon}_H$ & $\mathbf{\epsilon}_L$ & $\mathbf{\Delta\epsilon}$ & \tcap{ZPVE}       & $\mathbf{C}_v$       \\ \midrule
	GraphMVP\tsup{1}       & 0.031                & 0.070                & 28.5                  & 26.3                  & 46.9                      & 1.63              & 0.033                \\
	GEM\tsup{2}            & 0.034                & 0.081                & 33.8                  & 27.7                  & 52.1                      & 1.73              & 0.035                \\
	3D Infomax\tsup{3}     & 0.034                & 0.075                & 29.8                  & 25.7                  & 48.8                      & 1.67              & 0.033                \\
	3D-MGP\tsup{4}         & 0.020                & 0.057                & 21.3                  & 18.2                  & 37.1                      & 1.38              & 0.026                \\ \midrule
	PhysNet\tsup{5}        & 0.053                & 0.062                & 32.9                  & 24.7                  & 42.5                      & 1.39              & 0.028                \\
	Schnet\tsup{6}         & 0.033                & 0.235                & 41.0                  & 34.0                  & 63.0                      & 1.7               & 0.033                \\
	Cormorant\tsup{7}      & 0.038                & 0.085                & 34.0                  & 38.0                  & 61.0                      & 2.03              & 0.026                \\
	DimeNet++\tsup{8}      & 0.030                & 0.044                & 24.6                  & 19.5                  & 32.6                      & 1.21              & 0.023                \\
	PaiNN\tsup{9}          & \rgd{0.012}          & 0.045                & 27.6                  & 20.4                  & 45.7                      & 1.28              & 0.024                \\
	EGNN\tsup{10}          & 0.029                & 0.071                & 29.0                  & 25.0                  & 48.0                      & 1.55              & 0.031                \\
	NoisyNode\tsup{11}     & 0.025                & 0.052                & 20.4                  & 18.6                  & 28.6                      & \rgd{1.16}        & 0.025                \\
	SphereNet\tsup{12}     & 0.025                & 0.053                & 22.8                  & 18.9                  & 31.1                      & \rbs{1.12}        & 0.024                \\
	SEGNN\tsup{13}         & 0.023                & 0.060                & 24.0                  & 21.0                  & 42.0                      & 1.62              & 0.031                \\
	EQGAT\tsup{14}         & \rbs{0.011}          & 0.053                & 20.0                  & 16.0                  & 32.0                      & 2.00              & 0.024                \\\midrule
	SE(3)-T\tsup{15}       & 0.051                & 0.142                & 35.0                  & 33.0                  & 53.0                      & -                 & 0.052                \\
	TorchMD-Net\tsup{16}   & \rbs{0.011}          & 0.059                & 20.3                  & 17.5                  & 36.1                      & 1.84              & 0.026                \\
	Equiformer\tsup{17}    & \rbs{0.011}          & 0.046                & \rgd{15.0}            & \rgd{14.0}            & 30.0                      & 1.26              & 0.023                \\
	Transformer-M\tsup{18} & 0.037                & \rgd{0.041}          & 17.5                  & 16.2                  & \rgd{27.4}                & 1.18              & \rgd{0.022}          \\ \midrule
	TGT-Ag                 & 0.025                & \rbs{0.040}          & \rbs{9.9}             & \rbs{9.7}             & \rbs{17.4}                & 1.18              & \rbs{0.020}          \\
	\bottomrule
\end{tabular}

%% file: tables/molpcba_results.tex
\begin{tabular}{lcc}
	\toprule
	\textbf{Model}                                              & \textbf{\#Param} & \textbf{Test AP\hib{}(\%)}   \\ \midrule
	DeeperGCN\tsup{1}-VN\tsup{9}-FLAG\tsupsub{10}               & 6.55M            & 28.42\istd{0.43}       \\
	PNA\tsup{2}                                                 & 6.55M            & 28.38\istd{0.35}       \\
	DGN\tsup{3}                                                 & 6.73M            & 28.85\istd{0.30}       \\
	GINE\tsup{4}-VN\tsup{9}                                     & 6.15M            & 29.17\istd{0.15}       \\
	PHC-GNN\tsup{5}                                             & 1.69M            & 29.47\istd{0.26}       \\ \midrule
	GIN\tsup{6}-VN\tsupsub{9}{pretrain}                         & 3.4M             & 29.02\istd{0.17}       \\
	Graphormer\tsup{7}-FLAG\tsupsub{10}{pretrain}               & 119.5M           & \rgd{31.40\istd{0.34}} \\
	EGT\tsupsub{8}{pretrain}                                    & 110.8M           & 29.61\istd{0.24}       \\ \midrule
	EGT\tsup{8}+RDKit                                           & 47M              & \rgd{31.09\istd{0.33}} \\
	EGT\tsup{8}+TGT-At-DP                                       & 47M              & \rgd{31.12\istd{0.25}} \\
	TGT-Ag+RDKit                                                & 47M              & \rgd{31.44\istd{0.29}} \\
	TGT-Ag+TGT-At-DP                                            & 47M              & \rbs{31.67\istd{0.31}} \\
	\bottomrule
\end{tabular}

%% file: tables/litpcba_avg_results_trunc.tex
\begin{tabular}{lcc}
	\toprule
	                                      & \textbf{Avg. Test}   \\
	\textbf{Model}                        & \textbf{ROC-AUC\hib{}(\%)} \\ \midrule
	GCN\tsup{1}                           &      72.3            \\
	GAT\tsup{2}                           &      75.2            \\
	FP-GNN\tsup{3}                        &      75.9            \\ \midrule
	EGT\tsup{4}                           &      66.7            \\
	EGT\tsupsub{4}{pretrain}              &      78.9            \\
	GEM\tsup{5}                           &      76.6            \\
	GEM\tsupsub{5}{pretrain}              &      78.4            \\
	GEM-2\tsup{6}                         &      77.6            \\
	GEM-2\tsupsub{6}{pretrain}            & \rbs{81.5}           \\ \midrule
	EGT\tsup{4}+RDKit                     & \rgd{81.2}           \\
	EGT\tsup{4}+TGT-At-DP                 & \rbs{81.5}           \\
	\bottomrule
\end{tabular}

%% file: tables/tsp_results.tex
\begin{tabular}{lcc}
	\toprule
	                               & \textbf{Test F1\hib{}(\%)}        & \textbf{Test F1\hib{}(\%)}         \\
	\textbf{Model}                 & \textbf{(\#Param$\approx$100K)}  & \textbf{(\#Param$\approx$500K)}   \\\midrule
	GCN\tsup{1}                    & 63.0\istd{0.1}          & -                        \\
	GIN\tsup{2}                    & 65.6\istd{0.3}          & -                        \\
	GAT\tsup{3}                    & 67.1\istd{0.2}          & -                        \\
	GatedGCN\tsup{4}               & 80.8\istd{0.3}          & 83.8\istd{ 0.2}          \\
	Graphormer\tsup{5}             & -                       & 69.8\istd{ 0.7}          \\
	ARGNP\tsup{6}                  & -                       & 85.5\istd{ 0.1}          \\
	EGT\tsup{7}                    & 82.2\istd{0.0}          & 85.3\istd{ 0.1}          \\\midrule
	TGT-Ag                         & 83.2\istd{0.1}          & 85.7\istd{ 0.1}          \\
	TGT-Agx2                       & 84.9\istd{0.0}          & 86.2\istd{ 0.1}          \\
	TGT-Agx3                       & \rgd{85.2\istd{0.1}}    & \rgd{86.6\istd{ 0.1}}    \\
	TGT-Agx4                       & \rbs{85.4\istd{0.1}}    & \rbs{87.1\istd{ 0.1}}    \\
	TGT-At                         & 83.3\istd{0.1}          & -                        \\
	\bottomrule
\end{tabular}

%% file: tables/triplet_dp_compare.tex
\setlength{\tabcolsep}{2pt}
\begin{tabular}{lccccccc}
	\toprule
	                  & No        & Axial & Triang. & Ungated   & Triplet & Ungated     & Triplet     \\
	                  & 3rd-      & Att.  & Update  & Triplet   & Agg.    & Triplet     & Att.        \\
	                  & Order     &       &         & Agg.      &         & Att.        &             \\ \midrule
	Cross-Ent.\lob{}  & 1.270     & 1.231 & 1.225   & 1.226     & 1.218   & \rgd{1.207} & \rbs{1.199} \\
	Time/Epoch\lob{}  & \rbs{1.0} & 2.2   & 1.8     & \rgd{1.6} & 1.7     & 3.1         & 3.3         \\
	\bottomrule
\end{tabular}

%% file: tables/ablation.tex
\setlength{\tabcolsep}{3pt}
\begin{tabular}{ccccccccc}
	\toprule
	Stoch. & RDKit   & Denoise & Noise   & Source & DFT    & DFT    & Tripl. & Val.       \\
	Infer. & Coords. & Obj.    & Local   & Drop.  & Pre-   & Dist.  & Att.   & MAE\lob{}  \\
	       & Input   &         & Smooth. &        & train. & Pred.  &        & (meV)      \\ \midrule
	-      & -       & -       & -       & -      & -      & -      & -      & 85.1       \\
	\mtick & -       & -       & -       & -      & -      & -      & -      & 84.2       \\
	\mtick & \mtick  & -       & -       & -      & -      & -      & -      & 82.2       \\
	\mtick & \mtick  & \mtick  & -       & -      & -      & -      & -      & 80.9       \\
	\mtick & \mtick  & \mtick  & \mtick  & -      & -      & -      & -      & 80.5       \\
	\mtick & \mtick  & \mtick  & \mtick  & \mtick & -      & -      & -      & 80.1       \\
	\mtick & \mtick  & \mtick  & \mtick  & \mtick & -      & -      & \mtick & 79.4       \\
	\mtick & \mtick  & \mtick  & \mtick  & \mtick & \mtick & -      & -      & 75.3       \\
	\mtick & \mtick  & \mtick  & \mtick  & \mtick & -      & \mtick & -      & 76.6       \\
	\mtick & \mtick  & \mtick  & \mtick  & \mtick & \mtick & \mtick & -      & \rgd{72.9} \\
	\mtick & \mtick  & \mtick  & \mtick  & \mtick & \mtick & \mtick & \mtick & \rbs{71.0} \\
	\bottomrule
\end{tabular}

%% file: appendix.tex
\section{Additional Details about Related Works}
\label{sec:apx_related}
\textbf{Graph Transformers}
Before pure graph transformers, the self-attention mechanism was used to boost the expressivity of localized message-passing Graph Neural Networks (GNNs) - for example, GraphTrans \citep{wu2021representing} and GSA \citep{wang2021global} used global self-attention to improve long-range information exchange in GNNs. GROVER \citep{rong2020self} utilized GNNs to generate query, key, and value matrices for self-attention, enabling pretraining on molecular graphs. These hybrid approaches were followed by a new research interest in pure graph transformers. SAN \citep{kreuzer2021rethinking} utilized Laplace Positional Encodings (LPE) in a global self-attention based graph transformer. Graphormer \citep{ying2021transformers} proposed graph-specific relative positional encodings and showed superior performance on molecular property prediction tasks. EGT \citep{hussain2022global} extended the transformer framework to include pairwise/edge channels and proposed a general framework for graph learning including edge-related and pairwise tasks. GEM-2 \citep{liu2022gem} extended the notion of pairwise channels to include higher-order channels to account for many body interactions in molecular graphs. GRPE \citep{park2022grpe} proposed a more expressive relative positional encoding scheme for graphs. TokenGT \citep{kim2022pure} proposed to include both nodes and edges as tokens in the transformer. UniMol and UniMol+ \citep{lu2023highly} use transformer backbones with pairwise channels, similar to EGT, for molecular property prediction. GPS \citep{rampavsek2022recipe} proposed a framework to combine message-passing and self-attention mechanisms and GPS++ \citep{masters2022gps++} tuned these choices to achieve SOTA performance on PCQM4Mv2. Our TGT model is a pure transformer architecture for graph learning with novel triplet interaction mechanisms for 3rd-order interaction between neighboring pairs which is computationally much cheaper than higher-order channels used in GEM-2 while still being expressive enough to capture geometric information required for molecular property prediction.

\ssubsec{Molecular Property Prediction}
Following the success of message-passing GNNs on molecular property prediction tasks \citep{gilmer2017neural}, new geometry and physics informed GNNs have been proposed which are equivariant or invariant to 3D rotations and translations. Works like SchNet \citep{schutt2017schnet} and DimeNet \citep{gasteiger2020directional} use distance-based convolution whereas spherical methods like GemNet \citep{gasteiger2021gemnet} and SphereNet \citep{liu2021spherical} also encode angle information. Equivariant aggregation was later generalized to equivariant transformers like Equiformer \citep{tholke2021equivariant} and TorchMD-Net \citep{tholke2022torchmd}. Unlike these works which innovate on the network architecture to preserve equivariance, we preserve SE(3) invariance by limiting the input features to interatomic distances.

\ssubsec{3D Pretraining}
While ground truth 3D structural information, e.g., atomic coordinates optimized through density functional theory (DFT) improves model accuracy, they are prohibitively expensive to compute for each inference instance. 3D pretraining approaches address this by using 3D data sources to teach encoders useful structural knowledge. These pretrained networks can then effectively process 2D molecular graphs for property prediction where explicit 3D data is unavailable. For example, 3D Infomax \citep{stark20223d} and GraphMVP \citep{liu2021pre} maximize the mutual information between 2D topological and 3D views. Chemrl-GEM \citep{fang2021chemrl} uses bond-angle graphs and reconstruction tasks with approximate 3D data. 3D PGT \citep{wang2023automated} combines multiple generative tasks on 3D conformations guided by a total energy signal. GeomSSL \citep{liu2022molecular} proposes coordinate and distance denoising to model potential energy surfaces. Transformer-M \citep{luo2022one} encodes distances into self-attention and also trains the transformer to be able to predict in the absence of 3D information. UniMol+ \citep{lu2023highly} iteratively refines cheaply computed RDKit coordinates before making final predictions. In contrast to these methodologies, our approach involves training a distance predictor that directly forecasts interatomic distances from 2D molecular graphs. This eliminates the need for initial 3D coordinates in downstream tasks, as the predicted distances serve as direct inputs.

\section{Third-order Interactions in Previous Works}
\label{sec:apx_pair_to_pair}

\subsection{Axial Attention}
GEM-2 \citep{liu2022gem} introduced axial attention, which can be simplified when we consider \emph{only} pairwise channels as follows:
\begin{align}
    \mathbf{o}_{ij} &= \sum_{k=1}^N a_{ijk} \mathbf{v}_{jk} \label{eqn:axial_at1}\\
    a_{ijk} &= \softmax_k\left(\frac{1}{\sqrt{d}}\mathbf{q}_{ij}\cdot\mathbf{p}_{jk}\right) \label{eqn:axial_at2}
\end{align}
where $\mathbf{v}_{jk}$ is the value vector of the pair $(j,k)$ and $a_{ijk}$ is the attention weight between pairs $(i,j)$ and $(j,k)$ (we have re-used the notations from node-to-node attention for consistency). $\mathbf{q}_{ij}$ and $\mathbf{p}_{jk}$ are the query and key vectors of pairs $(i,j)$ and $(j,k)$, respectively. This is a generalization of the self-attention mechanism for pairs and has a computational complexity of $O(N^3)$. However, notice that the neighboring pair $(i,k)$ of the 3-tuple $(i,j,k)$ is not considered in this process. GEM-2 \citeGEMtwo{} instead uses a 3rd-order channel to provide positional information to this attention process. However, in the absence of 3rd-order information like angles, axial attention does not perform well. Another update is done by changing the direction of the aggregated pairs, i.e., from $(k, j)$ to $(i, j)$, with the weights $a_{ikj}$.

\subsection{Triangular Update}
The triangular update proposed by AlphaFold 2 \citep{jumper2021highly} and later adopted by UniMol+ \citep{lu2023highly} takes the form:
\begin{align}
    o_{ij} = \sum_{k=1}^N a_{ik}b_{jk} \label{eqn:tri_up1}
\end{align}
where, $a_{ik}$ and $b_{jk}$ are scalars formed from the projections of the pair embeddings $\mathbf{e}_{ik}$ and $\mathbf{e}_{jk}$, respectively, and the output $o_{ij}$ is also a scalar. However, this mechanism is done for multiple sets of projections and the outputs are concatenated. Additionally, another update takes place in the opposite direction, i.e., for $o'_{ij} = \sum_{k=1}^N a'_{ki}b'_{kj}$.

Notice that, in this case, the update is a simple matrix/tensor multiplication which can have subcubic complexity. However, the information flow from pair $(j,k)$ to pair $(i,j)$ is mediated only by the pair $(i,k)$. Unlike a true attention process, $(i,j)$ cannot ``select'' which pairs to aggregate. The information passed for each set is a scalar, which means that many sets are required compared to a few attention heads. Also, this summation/aggregation process is unbounded which can be problematic if the input graphs vary in size drastically.

\section{The Edge-augmented Graph Transformer (EGT)}
\label{sec:apx_egt}

The Edge-augmented Graph Transformer \citeEGT{} is an extension to the transformer framework by \citet{vaswani2017attention} for general-purpose graph learning. This architecture uses the embeddings $\mathbf{h}_i$, where $i \in \{1, \ldots, N\}$, to represent the nodes in a graph with $N$ nodes. The contribution of EGT is to add additional edge channels with $N \times N$ pairwise embeddings $\mathbf{e}_{ij}$ which represent both existing and non-existing edges. The edge channels (i.e., pairwise embeddings) are updated both in the multi-head attention and their own feed-forward layers, just like the node embeddings. In this way, EGT makes the graph topology dynamic and allows new pairwise representations to emerge over consecutive layers.

\ssubsec{EGT Multi-head Attention}
We adopt the EGT architecture with some changes. Firstly, we remove the dot product clipping in the multi-head attention layer, which was introduced as a means for stabilizing the training. With this change, the EGT self-attention mechanism can be expressed as:
\begin{align}
    \mathbf{o}_i = \sum_{j=1}^N a_{ij} \mathbf{v}_j
    \label{eq:egt_attention}
\end{align}
where $\mathbf{v}_j$ is the value vector of node $j$ and $a_{ij}$ is the attention weight between nodes $i$ and $j$. The attention weight is computed as:
\begin{align}
    a_{ij} &= \softmax_j(t_{ij})  \times \sigma(g_{ij})\\
    t_{ij} &= \frac{1}{\sqrt{d_k}} \mathbf{q}_i \cdot \mathbf{k}_j + b_{ij}
\end{align}
where $\mathbf{q}_i$ and $\mathbf{k}_j$ are the query and key vectors of nodes $i$ and $j$, respectively, $d_k$ is the dimension of the key vectors, and $\mathbf{o}_i$ is the output vector which is used to update the node embedding $\mathbf{h}_i$. The attention logit $t_{ij}$ is the summation of the scaled dot product (between the query and key vectors) and the attention bias. The edge channels participate in the attention mechanism by (i) providing a bias term $b_{ij}$ and (ii) providing a gating term $g_{ij}$ which passes through a sigmoid function $\sigma(\cdot)$ to directly control the flow of information from node $j$ to node $i$. Both the bias and gating terms are computed from projections of the edge embeddings $\mathbf{e}_{ij}$. This is done for each head of the multi-head attention mechanism in each layer. The node channels are updated from the projection of the concatenation of $o_i$ of all heads. The edge channels are also updated from a projection of the concatenation of the attention logits $t_{ij}$ of all heads. This way, EGT ensures two-way communication between the node and edge channels in the multi-head attention layer. This is in contrast to architectures like UniMol+ \citep{lu2023highly} where the edge channels are updated from an outer product of the node embeddings which adds additional computational cost.

\ssubsec{Dynamic Centrality Scalers}
We also adopt the dynamic centrality scalers introduced by EGT which ensures that the network is sensitive to the centralities of the nodes and thus at least 1-WL expressive. The centrality scalers are computed from the abovementioned gating terms $g_{ij}$ as:
\begin{align}
    \mathbf{s}_i = \ln \sum_{j=1}^N (1 + \sigma(g_{ij}))
\end{align}
which scales the output $o_i$ of each attention head. 

\ssubsec{Other Details}
Similar to current best practices for transformers, EGT uses a pre-norm layer normalization \citep{ba2016layer} before the multi-head attention layer and the feed-forward layer (FFN). EGT uses separate FFNs for the node and edge channels.

\ssubsec{Other Modifications}
Unlike the original EGT architecture which uses virtual nodes for graph-level tasks, we use a simple global average pooling over the final node embeddings to produce graph-level representations. Also, we do not use any form of absolute positional encoding, like the SVD-based positional encoding used by EGT.

\section{Additional Details for TGT Model}
\subsection{Source Dropout}
\begin{figure}[htbp]
    \centering
    \includegraphics[width=0.5\linewidth]{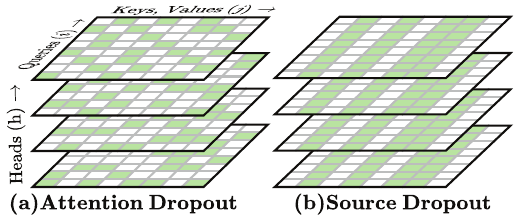}
    \caption{(a) Attention Dropout vs. (b) Source Dropout.}
    \label{fig:apx_source_dropout}
\end{figure}
\newcommand{\refFigSourceDropout}{\cref{fig:apx_source_dropout}}

Source dropout is an attention-masking process similar to attention dropout. For each sample in a batch of samples, we randomly mask the columns of the attention matrix by adding a large negative value. The value for column $j$ is $m_j = -\infty$ with probability $p$, and $m_j = 0$ with probability $1-p$. Then, the bias term $b_{ij}^h = m_j$ for all nodes $i$ and attention heads $h$, is added to the input of the softmax function. This is illustrated in \refFigSourceDropout{}. Notice that this pattern will be different for different layers and also for different samples in the batch. It is called ``source'' dropout because it essentially makes some of the nodes unavailable as information sources for all other nodes, in a particular layer. This is in contrast to attention dropout where the information may still flow via other attention heads.

\subsection{Distance Encoding}
We use two forms of encoding schemes to encode the continuous interatomic distances into the input features to the edge channels. The first one is an RBF encoding scheme, similar to the one used by Transformer-M \citep{luo2022one}. The second one is a Fourier encoding scheme. Both encodings perform well, and the choice can be made based on the application.

\ssubsec{RBF Encoding}
The RBF encoding scheme is defined as:
\begin{align}
o_{ij}^k = \frac{1}{\sqrt{2\pi} \cdot |\sigma^k|} \exp\left[ - \frac{1}{2} \left(\frac{m_{ij}^k \cdot d_{ij} + b_{ij}^k - \mu^k}{|\sigma^k|} \right)^2 \right]
\end{align}
Where $d_{ij}$ is the interatomic distance between atoms $i$ and $j$. $m_{ij}^k$, $b_{ij}^k$, $\mu^k$, and $\sigma^k$ are learnable parameters for the $k$-th kernel. $m_{ij}^k$ and $b_{ij}^k$ are looked up from an embedding table based on the type of atom pair $(i,j)$. This is done for multiple kernels, each with a different set of parameters. The output of the RBF encoding is the concatenation of the outputs of all kernels which is fed through a two-layer MLP to produce the final output.

\ssubsec{Fourier Encoding}
The Fourier encoding scheme is defined as:
\begin{align}
    \mathbf{o}_{ij}^k &= \left[ \sin(\phi_{ij}^k), \cos(\phi_{ij}^k) \right]\\
    \phi_{ij}^k &= d_{ij} \times \frac{2\pi}{\lambda_k}
\end{align}
Where, $[\cdot, \cdot]$ represents concatenation. $d_{ij}$ is the interatomic distance between atoms $i$ and $j$. $\lambda_k$ is the wavelength associated with the $k$-th kernel. $\phi_{ij}^k$ is the phase for the $k$-th kernel at the distance $d_{ij}$. The output of the Fourier encoding is the concatenation of the outputs of all kernels which is fed through a linear layer to produce the final output. We choose the wavelengths $\lambda_k$ to be logarithmically spaced between $2\times\delta_{min}$ and $2\times\delta_{max}$, where $\delta_{min}$ and $\delta_{max}$ are the minimum and maximum interatomic distances of interest, respectively.

\subsection{Feature Encoding}
For molecular data, we use the same set of atomic and bond features as provided by the OGB \citeOGB{} Python library. These features are transformed via a learnable vector embedding layer before being fed to the node and edge channels, respectively. Additionally, we use the shortest path hop encoding scheme of EGT \citeEGT{}, which is also transformed via a learnable vector embedding layer before being fed to the edge channels. For the OC20 dataset, we use only an embedding of atomic numbers as node features, and the distance encoding mentioned above as edge features.

\subsection{Parameter Sharing in Consecutive Layers}
We found that a few subsequent TGT layers can share the same set of parameters, similar to ALBERT \citep{lan2019albert}. More specifically, layers $\{i \times m + j +1\}$ for $j \in \{0, \ldots, m-1\}$ share the same set of parameters, where $m$ is the ``layer multiplier''. We refer to this as TGTx$m$.

This can be useful for reducing the number of parameters in the model, as a form of model compression. However, for a given compute budget, this does not significantly reduce the computational and memory costs of training or inference, and it is more efficient to use separate parameters for each layer. However, this can become more relevant as the model size increases by allowing the model to fit within the GPU memory. This can also be useful for communication-bound distributed training, as the gradients of the shared parameters are computed only once and then broadcast to all the layers. This form of compression can be useful for the storage of the model as well.

\section{Additional Details about Datasets and Training}
\label{sec:apx_more_training}

\subsection{PCQM4Mv2}
The PCQM4Mv2 dataset, comprising 4 million molecules, is a part of the OGB-LSC datasets \citeOGBLSC{}. The primary objective involves predicting the quantum chemical property known as the HOMO-LUMO gap, representing the energy difference between the Highest Occupied Molecular Orbital (HOMO) and the Lowest Unoccupied Molecular Orbital (LUMO). The molecular formulas are provided as SMILES strings. The 2D graph can be efficiently extracted using RDKit \citeRDKit{}, along with pertinent node (atom) and edge (bond) features. We employ the same feature set from the OGB-LSC Python library. The ground truth 3D positions of atoms, derived from DFT (Density Function Theory) simulations, are provided in the training dataset. However, inference must be executed without DFT coordinates and within a reasonable time limit (4 hours).

To provide the distance predictor with initial 3D information, we utilize RDKit \citeRDKit{} to extract 3D coordinates and apply MM Force Field Optimization \citeMMFF{}, as outlined in \citeGEMtwo{}. It is important to note that this step is optional for our method.

Due to the absence of Ground Truth 3D coordinates in the validation set, we randomly divide the training set into train-3D and validation-3D splits, with the latter containing 5\% of the training data. Hyperparameters of the distance predictor are fine-tuned by monitoring the average cross-entropy loss of binned distance prediction on the validation-3D split, which is found to be a good indicator of downstream performance. The MAE (Mean Absolute Error) loss is employed to pretrain and finetune the task predictor, with an additional secondary cross-entropy loss for predicting ground truth distance bins with a weight of $0.1$. The input noise level is adjusted by evaluating the finetuned performance on the validation set. For a given noise level, the MAE during pretraining serves as a good indicator of downstream performance. We train both a 24-layer distance predictor and a 24-layer task predictor with identical architecture, utilizing the Adam optimizer. The distance predictor undergoes training for 60,000 steps with a batch size of 1024, while the task predictor is trained for 300,000 steps with a batch size of 2048 and finetuned for an additional 30,000 steps. This entire process is completed in less than 2 days, utilizing 32 NVIDIA V100 GPUs for our most resource-intensive TGT-At model. This approximately corresponds to 32 A100 GPU-days, slightly less than the training time of UniMol+ \citeUniMolP{}, which takes 40 A100 GPU-days. We get very good results by using an average of 10 sample predictions during stochastic inference, but to obtain the best possible results we draw 50 samples.

Despite our models having a higher parameter count compared to the previous SOTA UniMol+, when combining parameters of the distance predictor and the gap predictor, it is crucial to recognize that direct parameter count comparisons can be misleading, especially for iterative models like UniMol+, where parameters are shared across iterations, contrasting with non-iterative models like ours. To illustrate this we train a TGT-Ag model where the consequent layers share the same set of parameters, dubbed as TGT-Agx2 reducing parameter count by half yet still outperforming UniMol+. However, we do not resort to this form of parameter sharing because although it makes the model parameter-efficient, it does not significantly reduce the computational and memory costs of training and inference. Instead, we focus on compute efficiency, i.e., to get the best possible result for a given compute budget.

The validation MAE exhibits a high correlation with the test MAE for this dataset. We refrain from reporting test-dev MAE for all models due to the unavailability of test-dev labels and each evaluation of test-dev MAE requiring a leaderboard submission.

\subsection{Open Catalyst 2020 IS2RE}
The Open Catalyst 2020 Challenge \citeOCtwenty{} is designed to develop and evaluate machine learning models for predicting the adsorption energy of molecules on catalyst surfaces. We focus on the IS2RE (Initial Structure to Relaxed Energy) task of this benchmark where the model is provided with an initial DFT conformation of the crystal and adsorbate system, which interact with each other to reach the relaxed structure, at which point the energy of the system is measured. We exclusively use the IS2RE dataset for training which contains $\approx 460$K catalyst-adsorbate pairs.

A few changes are required to adapt our model to this task compared to molecular property prediction tasks. First, there is no 2D graph structure available, instead, we use the initial interatomic distance to provide relative positional information to both the distance predictor and the task predictor. The distance predictor is trained to predict the interatomic distances in the relaxed structure. The task predictor is pretrained on the noisy relaxed structure and later finetuned on the predicted interatomic distances by the distance predictor. MAE loss and a weighted denoising loss are used both during pretraining and finetuning. Due to the repeating nature of the crystal, we adopt the repeat and crop-by-distance approach of Graphormer-3D \citeGraphormerthreeD{}. However, we limit the number of atoms to a maximum of 64 by randomly sampling crystal atoms based on the proximity to the initial position of the adsorbate atoms. We also found that the distance range of interest for this task is slightly larger -- 16\r{A} compared to 8\r{A} for molecular graphs and a Fourier distance embedding works better than RBF-based distance embedding.

We train a 24-layer distance predictor and a 14-layer task predictor. The distance predictor is trained for 30,000 steps and the task predictor is pretrained for 100,000 steps and finetuned for 12,000 steps. This procedure takes approximately 2 days on 32 NVIDIA V100 GPUs for TGT-At. This approximately corresponds to 32 A100 GPU-days, which is significantly lower than the 112 GPU-days used by UniMol+. This is because we use much smaller sized graphs compared to UniMol+ and also our training method is more efficient. A median of 50 sample predictions is used for each input. We compare our results with other direct methods, i.e., methods that do not resort to iterative relaxation or molecular dynamics, and as such, only use the IS2RE data provided by OC20.

\subsection{QM9}
To highlight the transfer of knowledge to related quantum chemical prediction tasks we take the pretrained task predictor from our PCQM4Mv2 experiment and finetune it on the QM9 \citeQMnine{} dataset. QM9 is a quantum chemistry benchmark consisting of 134k small organic molecules. The ground truth 3D coordinates are provided on this dataset which can be used during inference, so the distance predictor is not required. The task instead is to predict different Quantum Mechanical properties as accurately as possible from the given 3D graph. We report finetuning performance on a subset of 7 tasks from the 12 tasks in QM9, namely, dipole moment ($\mu$), isotropic polarizability ($\alpha$), HOMO ($\epsilon_{H}$), LUMO ($\epsilon_{L}$) energies and their difference ($\Delta\epsilon$), Zero Point Vibrational Energy (ZPVE) and Heat Capacity ($C_v$). The results are presented in terms of MAE. Energies are expressed in meV. We use the same dataset splitting strategy as Transformer-M \citeTransfoM{} to form validation and test splits consisting of 10,000 and 10,831 molecules, respectively. We use MAE loss (normalized by the standard deviation of the task) and the Adam optimizer to finetune the pretrained task predictor model.

\subsection{MOLPCBA and LIT-PCBA}
Since 3D geometric information is valuable for molecular property prediction, we use our pretrained distance predictor (without RDKit) to provide an estimation of interatomic distances to models on the MOLPCBA molecular property prediction and the LIT-PCBA \citeLITPCBA{} drug discovery benchmarks.

The MOLPCBA dataset is a part of the OGB \citeOGB{} graph-level datasets, comprising 437,929 molecules collected from MoleculeNet \citep{wu2018moleculenet}. The task is to predict the presence or absence of 128 binary properties.

LIT-PCBA is a dataset for the virtual screening of 15 protein targets. It contains 9780 active compounds (positive samples) that bind to the targets, as well as 407,839 inactive compounds (negative samples) selected from PubChem Bioassay data. Predicting whether candidate compounds will bind to a particular target can be framed as a binary classification task. Since some of the proteins have very few positive samples, we use the same 7 targets (with over 150 active compounds each) and dataset splitting strategy as GEM-2 \citeGEMtwo{}.

Note that, since these datasets do not have ground truth 3D information, we do not finetune the distance predictor on these datasets, but rather use it as a frozen feature extractor. We train the task-specific predictor model from scratch on these datasets, with the extracted distance estimations as inputs. We also use RDKit coordinates as a secondary target while training to regularize the model, but inference can be performed in the absence of RDKit coordinates. We also compare against RDKit conformations as a source of 3D information. In our comparative results, when using RDKit input distances, we use the locally smooth noising and distance prediction objective mentioned in \cref{sec:pretrain} to train the task predictor to get the best achievable performance and to make a fair comparison with the distance predictor. We use the same model hyperparameters for both cases.

\label{sec:apx_hyperparams}
\begin{table}[htbp]
    \centering
    \caption{Hyperparameters used for each dataset.}
    \scalebox{0.85}{
        \input{tables/appendix/hyperparams.tex}
    }
    \label{tab:hyperparams}
\end{table}
\section{Hyperparameters}
The hyperparameters used for each dataset are presented in \cref{tab:hyperparams}. For PCQM4Mv2 and OC20 we list the hyperparameters for both the distance and the task predictor models and both training and finetuning. For QM9, we only list the hyperparameters for finetuning. For MOLPCBA, LIT-PCBA, and TSP we only show the hyperparameters for training from scratch. The missing hyperparameters do not apply to the corresponding dataset or model. For QM9 no secondary distance denoising objective is used. For LIT-PCBA, 0 triplet interaction heads indicate that an EGT is used without any triplet interaction module. For TSP datasets we train two models with 4 and 16 layers for parameter budgets 100K and 500K, respectively, which otherwise use the same hyperparameters.

\section{Additional Results}
\ssubsec{OC20 IS2RE}
The breakdown of performance on the OC20 IS2RE validation and test results are presented in \cref{tab:apx_oc20_val} and \cref{tab:apx_oc20_test}, respectively over the dataset splits ID (In Domain) and OOD (Out Of Domain) Adsorbates, Catalyst and Both. Notice that, TGT-At and UniMol+ outperform all other models for all splits.
\begin{table}[htbp]
    \centering
    \caption{Breakdown OC20 IS2RE validation results.}
    \label{tab:apx_oc20_val}
    \scalebox{0.85}{
        \input{tables/appendix/oc20_val.tex}
    }
    \newline
    \scriptsize{\tsup{1}\tciteSchnet{}, \tsup{2}\tciteDimeNetPP{}, \tsup{3}\tciteGemNet{}, \tsup{4}\tciteSphereNet{}, \tsup{5}\tciteGNS{}, \tsup{6}\tciteGraphormerthreeD{}, \tsup{7}\tciteEquiformer{}, \tsup{8}\tciteDRFormer{}, \tsup{9}\tciteMoleformer{}, \tsup{10}\tciteUniMolP{}, \tsup{11}\tciteNN {} }
\end{table}
\begin{table}[htbp]
    \centering
    \caption{Breakdown OC20 IS2RE test results.}
    \label{tab:apx_oc20_test}
    \scalebox{0.85}{
        \input{tables/appendix/oc20_test.tex}
    }
    \newline
    \scriptsize{\tsup{1}\tciteSchnet{}, \tsup{2}\tciteDimeNetPP{}, \tsup{3}\tciteSphereNet{}, \tsup{4}\tciteGNS{}, \tsup{5}\tciteGraphormerthreeD{}, \tsup{6}\tciteEquiformer{}, \tsup{7}\tciteDRFormer{}, \tsup{8}\tciteMoleformer{}, \tsup{9}\tciteUniMolP{}, \tsup{10}\tciteNN {} }
\end{table}

\ssubsec{LIT-PCBA}
We also show a breakdown of the LIT-PCBA results for the individual protein targets in \cref{tab:apx_litpcba_val}. We also compare against traditional machine learning methods like Naive Bayes \citeNaiveBayes{}, Support Vector Machine (SVM) \citeSVM{}, Random Forest (RF) \citeRndForest{}, and XGBoost \citeXGBoost{}. Notice that, EGT with our distance predictor TGT-At-DP outperforms other models in most cases except for GBA and MAPK1. We think this is due to the very low number of positive samples for these targets which is detrimental to training from scratch. We also see that the performance of TGT-At-DP is generally better than RDKit coordinates, which is a good indicator of the quality of the distance predictor.

\begin{table}[htbp]
    \centering
    \caption{LIT-PCBA results in terms of ROC-AUC\hib{} (\%).}
    \label{tab:apx_litpcba_val}
    \scalebox{0.85}{
        \input{tables/appendix/litpcba_results.tex}
    }
    \newline
    \scriptsize{\tsup{1}\tciteNaiveBayes{}, \tsup{2}\tciteSVM{}, \tsup{3}\tciteRndForest{}, \tsup{4}\tciteXGBoost{}, \tsup{5}\tciteGCN{}, \tsup{6}\tciteGAT{}, \tsup{7}\tciteFPGNN{}, \tsup{8}\tciteEGT{}, \tsup{9}\tciteGEM{}, \tsup{10}\tciteGEMtwo{}}
\end{table}

\begin{figure}[htbp]
    \centering
    \includegraphics[width=0.8\linewidth]{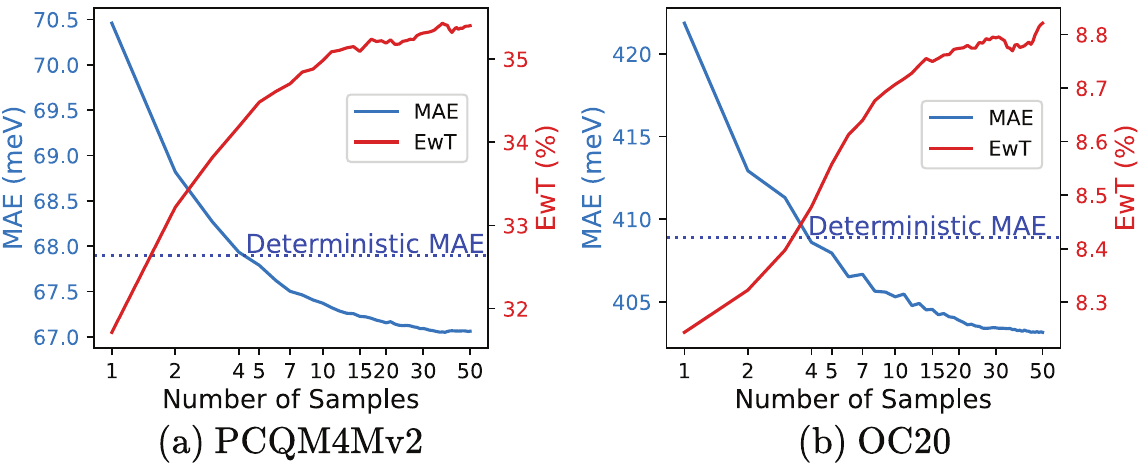}
    \caption{Number of samples drawn vs MAE(meV) and EwT(\%) on (a) PCQM4Mv2 and (b) OC20 IS2RE validation sets.}
    \label{fig:apx_samples}
\end{figure}

\section{Stochastic Inference Results}
To verify the cost and effectiveness of our proposed stochastic inference method, we illustrate how the performance improves with the number of stochastic samples drawn. We also evaluate the performance in a ``deterministic'' mode where the dropouts are turned off and only a single prediction is made (beforehand, we perform slight finetuning with dropout turned off for better performance). The results are presented in \cref{fig:apx_samples}. We see that the performance steadily improves with the number of samples drawn. It only takes 4-5 samples to outperform the deterministic prediction while with $\approx10$ samples we get very good results. The results continue to improve monotonically with more samples and approximately plateaus at $\approx50$ samples. Since these samples can be drawn in parallel and independently, with the performance exceeding the deterministic prediction by a fair margin, with only 10 samples they are a good trade-off between performance and cost.

\section{Distribution of Predictions}
\begin{figure}[htbp]
    \centering
    \includegraphics[width=0.8\linewidth]{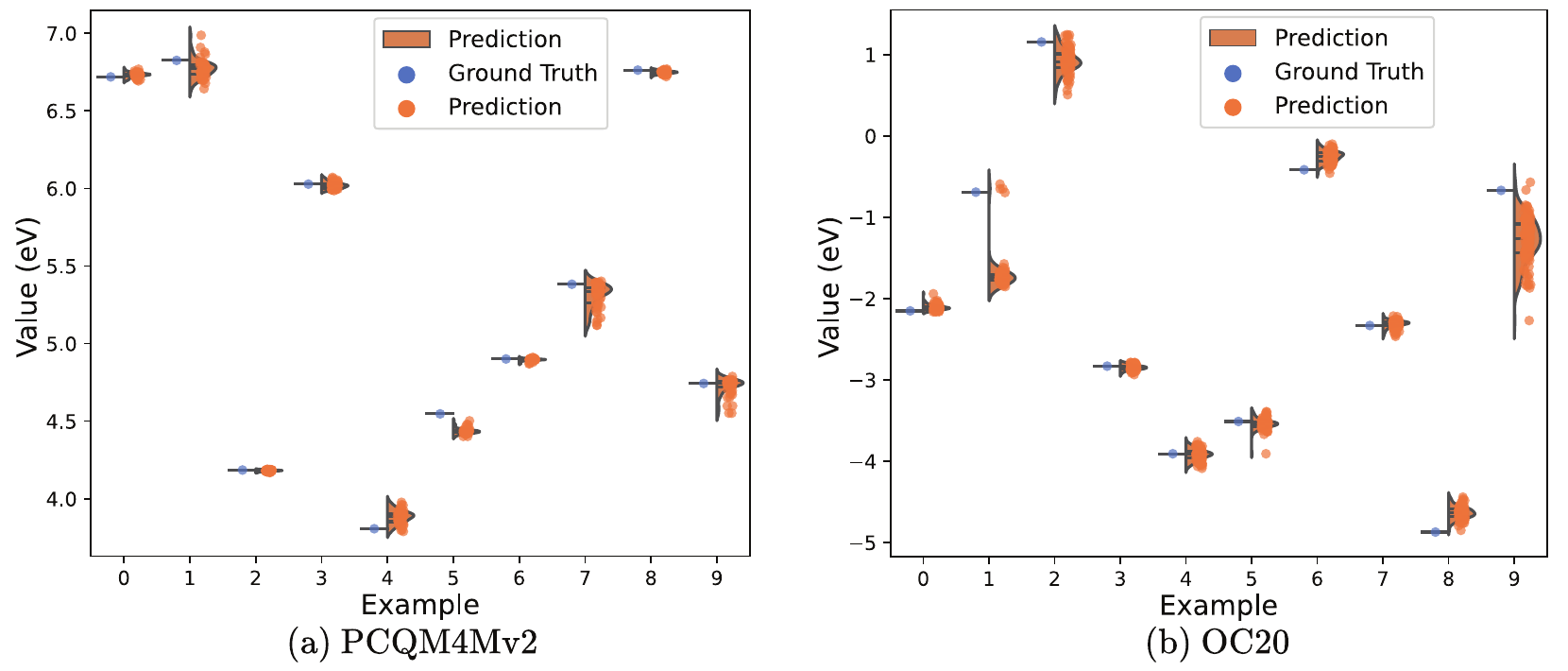}
    \caption{Example distributions of predictions vs. ground truth values for (a) PCQM4Mv2 and (b) OC20.}
    \label{fig:apx_distrib}
\end{figure}

In \cref{fig:apx_distrib} we show some example distributions of predictions vs. ground truth values for PCQM4Mv2 and OC20 IS2RE. We see that the predictions are generally centered around the ground truth values but for some cases, they can be multimodal. In these cases, the ground truth often corresponds to one of the modes, mostly the strongest mode.

Since we have multiple predictions, this raises the question of which statistic, i.e., mean, median, or mode, to use to produce a final prediction. In \cref{tab:apx_pcqm4m_stat}, and \cref{tab:apx_oc20_val_stat} and \cref{tab:apx_oc20_test_stat}, we show the performance of each statistic on the validation sets of PCQM4Mv2 and OC20 IS2RE, respectively. We see that for all cases the mean produces the best MAE but the worst EwT, whereas the mode produces the best EwT but also the worst MAE. The median is a good trade-off between the two. This indicates that the mean reduces the average error, while the mode improves the accuracy of the strongest predictions. The median is a robust statistic that is less sensitive to outliers and thus produces a good balance between the two. The choice of statistic can be made based on the application.

\begin{table}[htbp]
    \centering
    \caption{PCQM4Mv2 validation results for different statistics.}
    \label{tab:apx_pcqm4m_stat}
    \scalebox{0.85}{
        \input{tables/appendix/pcqm4m_stat.tex}
    }
\end{table}

\begin{table}[htbp]
    \centering
    \caption{OC20 validation results for different statistics.}
    \label{tab:apx_oc20_val_stat}
    \scalebox{0.85}{
        \input{tables/appendix/oc20_val_stat.tex}
    }
\end{table}

\begin{table}[htbp]
    \centering
    \caption{OC20 test results for different statistics.}
    \label{tab:apx_oc20_test_stat}
    \scalebox{0.85}{
        \input{tables/appendix/oc20_test_stat.tex}
    }
\end{table}

\section{Input 3D Noise and Local Smoothing}

\begin{figure}[htbp]
    \centering
    \includegraphics[width=0.8\linewidth]{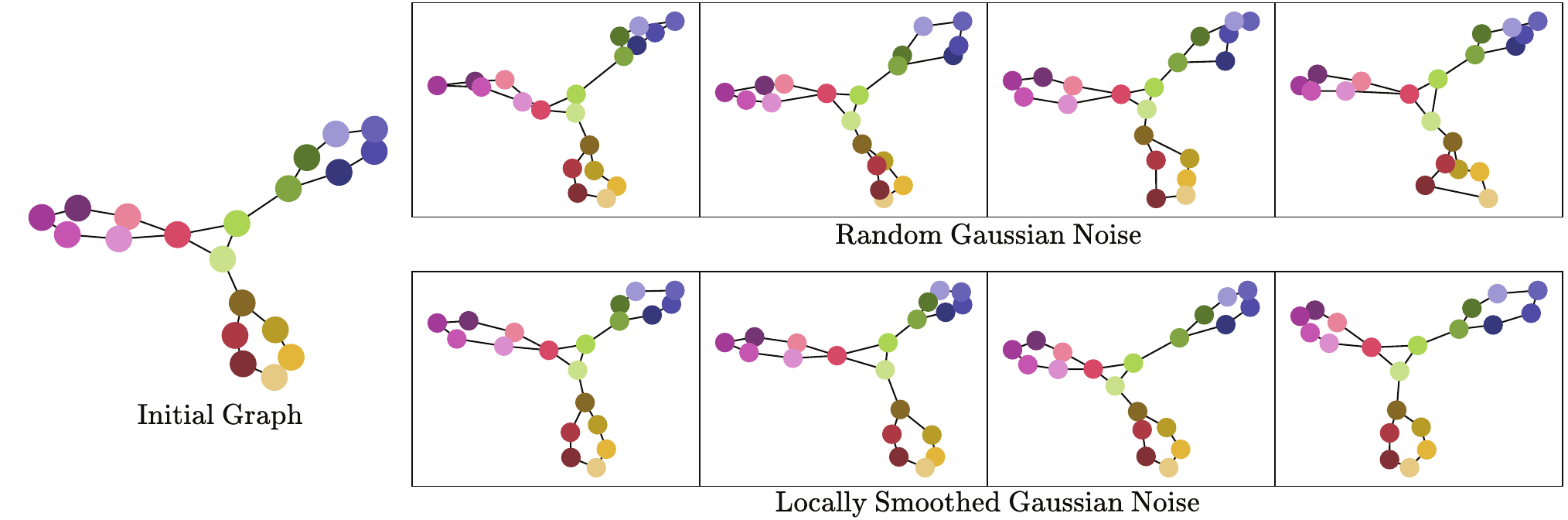}
    \caption{Effect of local smoothing on the injected noise for an example 2D graph.}
    \label{fig:apx_local_smooth}
\end{figure}

Since it is difficult to visualize the effect of input noise in 3D, we show its effect on an example 2D graph in \cref{fig:apx_local_smooth}. We see that, without local smoothing (i.e., random), the noise disproportionately affects the local structure of the graph. This is also in contrast to reality, where larger distances are more likely to be noisy/erroneous than smaller distances. With local smoothing, the noise mostly preserves the local structure of the graph, i.e., the nodes that are close together, also move together, whereas atoms that are far apart, move independently. This is more realistic and also encourages the model to utilize the local structure of the graph to make predictions.

\begin{figure}[htbp]
    \centering
    \includegraphics[width=0.5\linewidth]{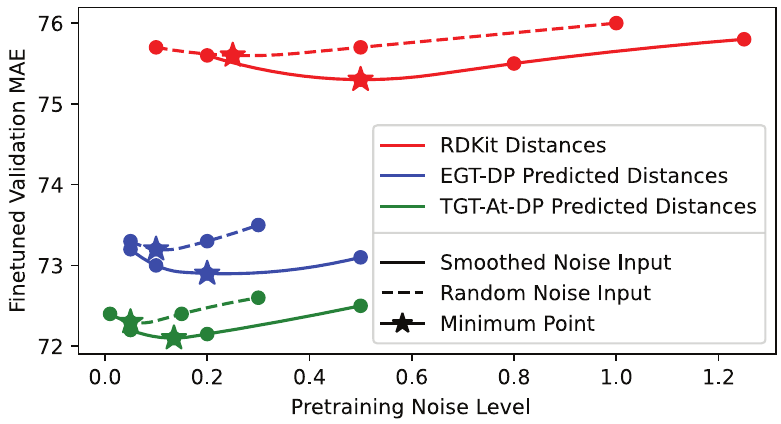}
    \caption{Pretraining input 3D noise vs finetuned performance for both random noise and locally smoothed noise, on the PCQM4Mv2 validation set for different downstream distance inputs.}
    \label{fig:apx_noise_curves}
\end{figure}

In \cref{fig:apx_noise_curves} we show the effect of input 3D noise on the finetuned performance on the PCQM4Mv2 validation set for both random noise and locally smoothed noise and different downstream distance inputs, i.e., interatomic distances from RDKit, an EGT distance predictor (EGT-DP) and our TGT-At distance predictor (TGT-At-DP). We see that, as the downstream distance input becomes more accurate (RDKit worst, EGT-DP better, TGT-At-DP best), the optimal noise level decreases. In all cases, better performance is achieved at the optimal noise level with local smoothing compared to random noise. Also, local smoothing allows us to inject more noise without degrading performance. This is because, without local smoothing, the higher level of noise perturbs the local structure of the molecule too much.


\section{Distribution of Pairwise Atomic Distances}
\begin{figure}[htbp]
    \centering
    \includegraphics[width=0.5\linewidth]{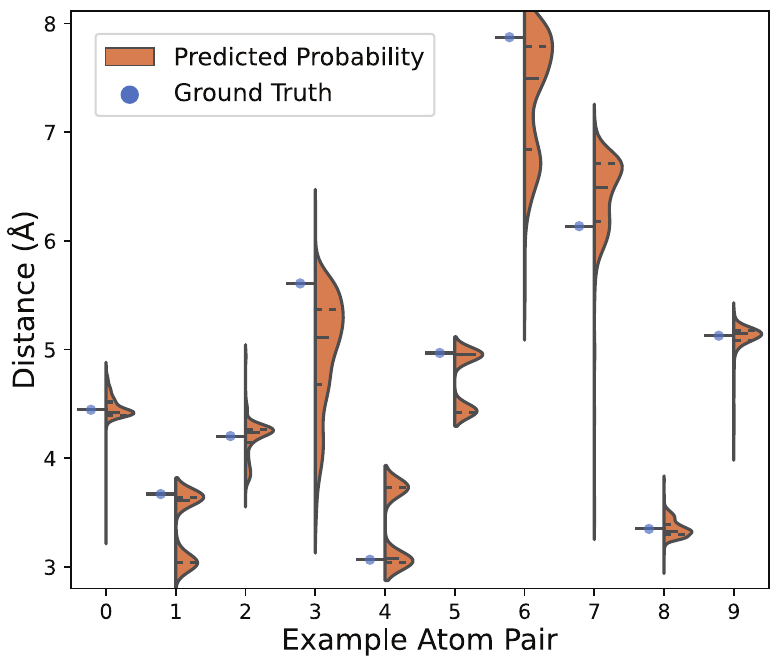}
    \caption{Example predicted distributions of pairwise atomic distances vs. ground truth values (from the PCQM4Mv2 dataset).}
    \label{fig:apx_atomic_dist}
\end{figure}

In \cref{fig:apx_atomic_dist} we show some example distributions of pairwise atomic distances predicted by the distance predictor, vs. ground truth values for the PCQM4Mv2 dataset. We see that for non-trivial distances, generally between non-bonded atoms, the predictions are often multimodal with multiple strong modes. Also, the distributions are often skewed. The ground truth distance is most likely to correspond to one of the modes, usually the strongest one. This is why it is important to use cross-entropy loss for the distance predictor, which encourages the model to learn the whole distribution. Using MSE or MAE loss would not encourage the model to learn the different modes of distribution.

\begin{table}[htbp]
    \centering
    \caption{Accuracy of pairwise distances in terms of MAE\lob{}, RMSE\lob{} and percent error within a threshold (EwT\hib{}).}
    \label{tab:apx_dist_acc}
    \scalebox{0.85}{
        \input{tables/appendix/dist_acc.tex}
    }
\end{table}

\section{How Accurate is the Distance Predictor?}
The distance predictor operates in the stochastic mode and outputs clipped and binned distances, which provides sufficient structural information to the downstream task predictor. However, it is difficult to directly compare the accuracy of the distance predictor to that of other methods like RDKit. To compare the accuracy of the distances we must first convert them into continuous unbounded distances. We do this by training a smaller refiner network which takes the clipped and binned distances as input and outputs continuous unbounded distances. We train this network with MAE loss and during stochastic inference take the median of the output distances. We compare the accuracy of individual pairwise distances on the validation-3D split of the PCQM4Mv2 dataset (i.e., unseen data during training), in terms of MAE, RMSE (Root Mean Square Error), and percent error within a threshold of 0.2\r{A}, 0.1\r{A}, 0.05\r{A} and 0.01\r{A} in \cref{tab:apx_dist_acc}.

We see that our TGT-At distance predictor outperforms RDKit by a large margin in terms of all of the metrics, which improves slightly if we feed RDKit distances as an initial estimate to the distance predictor. This is also reflected in the accuracy of the downstream task predictor. This indicates that the distance predictor can predict the underlying structure of the molecule more accurately than RDKit.

%% file: tables/appendix/hyperparams.tex
\begin{tabular}{l|cc|cc|c|c|c|c}
	\toprule
	                     & \multicolumn{2}{c|}{PCQM4Mv2} & \multicolumn{2}{c|}{OC20} & QM9         & MOLPCBA    & LIT-PCBA          & TSP                                           \\ \cmidrule(lr){2-3} \cmidrule(lr){4-5}
	Hyperparam.          & Dist. Pred.                   & Task Pred.                & Dist. Pred. & Task Pred. & Task Pred.       & Task Pred.       & Task Pred.       & -       \\ \midrule
	\# Layers            & 24                            & 24                        & 24          & 14         & 24               & 12               & 8                & 4,16    \\
	Node Embed. Dim.     & 768                           & 768                       & 768         & 768        & 768              & 768              & 1024             & 64      \\
	Edge Embed. Dim.     & 256                           & 256                       & 256         & 512        & 256              & 32               & 256              & 8       \\
	\# Attn. Heads       & 64                            & 64                        & 64          & 64         & 64               & 32               & 64               & 8       \\
	\# Triplet Heads     & 16                            & 16                        & 16          & 16         & 16               & 4                & 0                & 2       \\
	Node FFN Dim.        & 768                           & 768                       & 1536        & 768        & 768              & 768              & 2048             & 128     \\
	Edge FFN Dim.        & 256                           & 256                       & 512         & 512        & 256              & 32               & 512              & 16      \\
	Max. Hops Enc.       & 32                            & 32                        & -           & -          & 32               & 32               & 32               & 16      \\
	Activation           & GELU                          & GELU                      & GELU        & GELU       & GELU             & GELU             & GELU             & GELU    \\
	Input Dist. Enc.     & RBF                           & RBF                       & Fourier     & Fourier    & RBF              & RBF              & RBF              & Fourier \\ \midrule
	Source Dropout       & 0.3                           & 0.3                       & 0.3         & 0.3        & 0.3              & 0.3              & 0.3              & 0.1     \\
	Triplet Dropout      & 0.0                           & 0.0                       & 0.1         & 0.0        & 0.0              & 0.1              & 0.0              & 0.0     \\
	Path Dropout         & 0.2                           & 0.2                       & 0.2         & 0.1        & 0.2              & 0.1              & 0.1              & 0.0     \\
	Node Activ. Dropout  & 0.1                           & 0.1                       & 0.1         & 0.1        & 0.1              & 0.1              & 0.1              & 0.1     \\
	Edge Activ. Dropout  & 0.1                           & 0.1                       & 0.1         & 0.1        & 0.1              & 0.1              & 0.1              & 0.1     \\
	Input 3D Noise       & -                             & 0.2                       & -           & 0.6        & 0.0              & -                & -                & -       \\
	Input Noise Smooth.  & -                             & 1.0                       & -           & 1.0        & 0.0              & -                & -                & -       \\ \midrule
	Optimizer            & Adam                          & Adam                      & Adam        & Adam       & Adam             & Adam             & Adam             & Adam    \\
	Batch Size           & 1024                          & 2048                      & 256         & 256        & -                & 256              & 1024             & 32      \\
	Max. LR              & 0.001                         & 0.002                     & 0.001       & 0.001      & -                & $4\times10^{-4}$ & $5\times10^{-4}$ & 0.001   \\
	Min. LR              & $10^{-6}$                     & $10^{-6}$                 & 0.001       & $10^{-6}$  & -                & $10^{-6}$        & $5\times10^{-5}$ & $10^{-4}$  \\
	Warmup Steps         & 30000                         & 15000                     & 8000        & 16000      & -                & 4500             & 600              & 1000    \\
	Total Training Steps & 60000                         & 300000                    & 30000       & 100000     & -                & 30000            & 1200             & 20000   \\
	Grad. Clip. Norm     & 5.0                           & 5.0                       & 5.0         & 5.0        & 5.0              & 5.0              & 2.0              & 5.0     \\ \midrule
	Dist. Loss Weight    & -                             & 0.1                       & -           & 3.0        & 0.0              & 0.05             & 0.1              & -       \\
	\# Dist. Bins        & 256                           & 512                       & 256         & 512        & -                & 512              & 512              & -       \\
	Dist. Bins Range     & 8                             & 8                         & 16          & 16         & -                & 8                & 8                & -       \\ \midrule
	FT Batch Size        & -                             & 2048                      & -           & 1024       & 2048             & -                & -                & -       \\
	FT Warmup Steps      & -                             & 3000                      & -           & 0          & 3000             & -                & -                & -       \\
	FT Total Steps       & -                             & 50000                     & -           & 12000      & 150000           & -                & -                & -       \\
	FT Max. LR           & -                             & $2\times10^{-4}$          & -           & $10^{-5}$  & $2\times10^{-4}$ & -                & -                & -       \\
	FT Min. LR           & -                             & $10^{-6}$                 & -           & $10^{-5}$  & $10^{-6}$        & -                & -                & -       \\
	FT Dist. Loss Weight & -                             & 0.1                       & -           & 2.0        & 0.1              & -                & -                & -       \\
	\bottomrule
\end{tabular}

%% file: tables/appendix/oc20_val.tex
\begin{tabular}{l|ccccc|ccccc}
	\toprule
	                                 & \multicolumn{5}{c|}{MAE \lob{} (meV)} & \multicolumn{4}{c}{EwT \hib{} (\%)}                                                                                                             \\
	Model                            & ID                                    & OOD Ads.                            & OOD Cat.    & OOD Both    & Avg.        & ID         & OOD Ads.   & OOD Cat.    & OOD Both   & Avg.       \\ \midrule
	SchNet\tsup{1}                   & 646.5                                 & 707.4                               & 647.5       & 662.6       & 666.0       & 2.96       & 2.22       & 3.03        & 2.38       & 2.65       \\
	DimeNet++\tsup{2}                & 563.6                                 & 712.7                               & 561.2       & 649.2       & 621.7       & 4.25       & 2.48       & 4.40        & 2.56       & 3.42       \\
	GemNet-T\tsup{3}                 & 556.1                                 & 734.2                               & 565.9       & 696.4       & 638.2       & 4.51       & 2.24       & 4.37        & 2.38       & 3.38       \\
	SphereNet\tsup{4}                & 563.2                                 & 668.2                               & 559.0       & 619.0       & 602.4       & 4.56       & 2.70       & 4.59        & 2.70       & 3.64       \\
	GNS\tsup{5}                      & 540.0                                 & 650.0                               & 550.0       & 590.0       & 582.5       & -          & -          & -           & -          & -          \\
	GNS\tsup{5}+NN\tsup{11}          & 470.0                                 & 510.0                               & 480.0       & 460.0       & 480.0       & -          & -          & -           & -          & -          \\ \midrule
	Graphormer-3D\tsup{6}            & 432.9                                 & 585.0                               & 444.1       & 529.9       & 498.0       & -          & -          & -           & -          & -          \\
	EquiFormer\tsup{7}               & 422.2                                 & 542.0                               & 423.1       & 475.4       & 465.7       & 7.23       & 3.77       & 7.13        & 4.10       & 5.56       \\
	EquiFormer\tsup{7}+NN\tsup{11}   & 415.6                                 & 497.6                               & 416.5       & 434.4       & 441.0       & 7.47       & 4.64       & 7.19        & 4.84       & 6.04       \\
	DRFormer\tsup{8}                 & 418.7                                 & 486.3                               & 432.1       & 433.2       & 442.5       & 8.39       & 5.42       & 8.12        & 5.44       & 6.84       \\
	Moleformer\tsup{9}               & 413.0                                 & 523.0                               & 432.0       & 473.0       & 460.0       & 8.01       & 3.04       & 7.66        & 3.19       & 5.48       \\
	Uni-Mol+\tsup{10}                & \rbs{379.5}                           & \rgd{452.6}                         & 401.1       & \rgd{402.1} & \rgd{408.8} & \rbs{11.1} & \rgd{6.71} & 9.90        & 6.68       & \rgd{8.61} \\ \midrule
	TGT-Ag                           & 386.1                                 & 485.8                               & \rgd{394.8} & 428.1       & 423.7       & \rgd{10.8} & 6.55       & \rgd{10.27} & \rbs{6.92} & \rgd{8.64} \\
	TGT-At                           & \rgd{381.3}                           & \rbs{445.4}                         & \rbs{391.7} & \rbs{393.6} & \rbs{403.0} & \rbs{11.1} & \rbs{6.87} & \rbs{10.47} & \rgd{6.80} & \rbs{8.82} \\
	\bottomrule
\end{tabular}

%% file: tables/appendix/oc20_test.tex
\begin{tabular}{l|ccccc|ccccc}
	\toprule
	                                     & \multicolumn{5}{c|}{MAE \lob{} (meV)}                      & \multicolumn{4}{c}{EwT \hib{} (\%)}                                                      \\
	Model                                & ID            & OOD Ads.       & OOD Cat.    & OOD Both    & Avg.        & ID          & OOD Ads.    & OOD Cat.    & OOD Both    & Avg.        \\ \midrule
	SchNet\tsup{1}                       & 639.0         & 734.0          & 662.0       & 704.0       & 684.8       & 2.96        & 2.33        & 2.94        & 2.21        & 2.61        \\
	DimeNet++\tsup{2}                    & 562.0         & 725.0          & 576.0       & 661.0       & 631.0       & 4.25        & 2.07        & 4.1         & 2.41        & 3.21        \\
	SphereNet\tsup{3}                    & 563.0         & 703.0          & 571.0       & 638.0       & 618.8       & 4.47        & 2.29        & 4.09        & 2.41        & 3.32        \\
	GNS\tsup{4}+NN\tsup{10}              & 421.9         & 567.8          & 436.6       & 465.1       & 472.8       & 9.12        & 4.25        & 8.01        & 4.64        & 6.51        \\ \midrule
	Graphormer-3D\tsup{5}                & 397.6         & 571.9          & 416.6       & 502.9       & 472.2       & 8.97        & 3.45        & 8.18        & 3.79        & 6.10        \\
	EquiFormer\tsup{6}                   & 503.7         & 688.1          & 521.3       & 630.1       & 585.8       & 5.14        & 2.41        & 4.67        & 2.69        & 3.73        \\
	EquiFormer\tsup{6}+NN\tsup{10}       & 417.1         & 547.9          & 424.8       & 474.1       & 466.0       & 7.71        & 3.70        & 7.15        & 4.07        & 5.66        \\
	DRFormer\tsup{7}                     & 386.5         & 543.5          & 406.0       & 467.7       & 450.9       & 9.18        & 4.01        & 8.39        & 4.33        & 6.48        \\
	Moleformer\tsup{8}                   & 413.4         & 534.6          & 428.0       & 458.1       & 458.5       & 8.79        & 4.67        & 7.58        & 4.87        & 6.48        \\
	Uni-Mol+\tsup{9}                     & \rbs{374.5}   & \rgd{476.0}    & \rbs{398.0} & \rgd{408.6} & \rbs{414.3} & \rgd{11.29} & \rbs{6.05 } & \rgd{9.53 } & \rgd{6.06 } & \rgd{8.23 } \\ \midrule
	TGT-At                               & \rgd{379.6}   & \rbs{471.8}    & \rgd{399.0} & \rbs{408.4} & \rgd{414.7} & \rbs{11.50} & \rgd{5.70 } & \rbs{9.84 } & \rbs{6.17 } & \rbs{8.30 } \\
	\bottomrule
\end{tabular}

%% file: tables/appendix/litpcba_results.tex
\begin{tabular}{lcccccccc}
	\toprule
	                                              & ALDH1                & FEN1                 & GBA                & KAT2A              & MAPK1              & PKM2                 & VDR                  & Average    \\ \midrule
	No. active                                    & 7,168                & 369                  & 166                & 194                & 308                & 546                  & 884                  &            \\ 
	No. inactive                                  & 137,965              & 355,402              & 296,052            & 348,548            & 62,629             & 245,523              & 355,388              &            \\ \midrule
	NaiveBayes\tsup{1}                            & 69.3                 & 87.6                 & 70.9               & 65.9               & 68.6               & 68.4                 & 80.4                 & 73.0       \\
	SVM\tsup{2}                                   & 76.0                 & 87.7                 & 77.8               & 61.2               & 66.5               & 75.3                 & 69.                  & 73.4       \\
	RandomForest\tsup{3}                          & 74.1                 & 65.7                 & 59.9               & 53.7               & 57.9               & 58.1                 & 64.4                 & 62.0       \\
	XGBoost\tsup{4}                               & 75.0                 & 88.8                 & 83.0               & 50.0               & 59.3               & 73.7                 & 78.2                 & 72.6       \\ \midrule
	GCN\tsup{5}                                   & 73.0                 & 89.7                 & 73.5               & 62.1               & 66.8               & 63.6                 & 77.3                 & 72.3       \\
	GAT\tsup{6}                                   & 73.9                 & 88.8                 & 77.6               & 66.2               & 69.7               & 72.4                 & 78.0                 & 75.2       \\
	FP-GNN\tsup{7}                                & 76.6                 & 88.9                 & 75.1               & 63.2               & \rbs{77.1}         & 73.2                 & 77.4                 & 75.9       \\ \midrule
	EGT\tsup{8}                                   & 72.5\istd{1}         & 81.0\istd{5}         & 52.9\istd{12}      & 54.6\istd{1}       & 67.5\istd{2}       & 64.6\istd{4}         & 74.0\istd{1}         & 66.7       \\
	EGT\tsupsub{8}{pretrain}                      & 78.7\istd{2}         & 92.9\istd{1}         & 75.4\istd{4}       & 72.8\istd{1}       & \rgd{75.3\istd{3}} & 76.5\istd{2}         & 80.7\istd{2}         & 78.9       \\
	GEM\tsup{9}                                   & 77.6\istd{0.3}       & 93.3\istd{1}         & 82.9\istd{1}       & 63.2\istd{9}       & 68.5\istd{2}       & 73.5\istd{4}         & 77.1\istd{2}         & 76.6       \\
	GEM\tsupsub{9}{pretrain}                      & 77.2\istd{1}         & 91.4\istd{2}         & 82.1\istd{2}       & 74.0\istd{1}       & 71.0\istd{2}       & 74.6\istd{2}         & 78.5\istd{1}         & 78.4       \\
	GEM-2\tsup{10}                                & 77.0\istd{2}         & 92.9\istd{1}         & 81.9\istd{2}       & 67.0\istd{2}       & 71.5\istd{2}       & 72.4\istd{3}         & 80.5\istd{2}         & 77.6       \\
	GEM-2\tsupsub{10}{pretrain}                   & \rgd{80.2\istd{0.2}} & 94.5\istd{0.3}       & \rbs{85.6\istd{2}} & \rgd{76.3\istd{1}} & 73.3\istd{1}       & \rgd{78.2\istd{0.4}} & 82.3\istd{0.5}       & \rbs{81.5} \\\midrule
	EGT\tsup{8}+RDKit                             & \rgd{80.2\istd{0.2}} & \rgd{95.2\istd{0.3}} & \rgd{84.5\istd{4}} & 74.3\istd{1}       & 73.5\istd{1}       & 78.0\istd{0.2}       & \rgd{82.8\istd{0.3}} & \rgd{81.2} \\
	EGT\tsup{8}+TGT-At-DP                         & \rbs{80.6\istd{0.3}} & \rbs{95.5\istd{0.3}} & 84.4\istd{3}       & \rbs{74.6\istd{2}} & 74.3\istd{0.7}     & \rbs{78.4\istd{0.2}} & \rbs{82.9\istd{0.3}} & \rbs{81.5} \\
	\bottomrule
\end{tabular}

%% file: tables/appendix/pcqm4m_stat.tex
\begin{tabular}{lcc}
	\toprule
	Statistic & Val. MAE\lob{} (meV) & Val. EwT\hib{} (\%) \\ \midrule
	Mean      & \rbs{67.06}          & 35.40               \\
	Median    & \rgd{67.14}          & \rgd{36.08}         \\
	Mode      & 67.48                & \rbs{36.32}         \\
	\bottomrule
\end{tabular}

%% file: tables/appendix/oc20_val_stat.tex
\begin{tabular}{l|ccccc|ccccc}
	\toprule
	        & \multicolumn{5}{c|}{MAE\lob{} (meV)} & \multicolumn{4}{c}{EwT\hib{} (\%)}                                                                                                              \\
	Stastic & ID                                   & OOD Ads.                           & OOD Cat.    & OOD Both    & Avg.        & ID          & OOD Ads.   & OOD Cat.    & OOD Both   & Avg.       \\ \midrule
	Mean    & \rbs{380.5}                          & \rbs{444.2}                        & \rbs{391.1} & \rbs{392.8} & \rbs{402.2} & 10.68       & 6.63       & 10.20       & 6.41       & 8.48       \\
	Median  & \rgd{381.3}                          & \rgd{445.4}                        & \rgd{391.7} & \rgd{393.6} & \rgd{403.0} & \rgd{11.15} & \rgd{6.87} & \rgd{10.47} & \rgd{6.80} & \rgd{8.82} \\
	Mode    & 385.1                                & 449.0                              & 396.1       & 396.2       & 406.6       & \rbs{11.30} & \rbs{6.98} & \rbs{10.48} & \rbs{6.88} & \rbs{8.91} \\
	\bottomrule
\end{tabular}

%% file: tables/appendix/oc20_test_stat.tex
\begin{tabular}{l|ccccc|ccccc}
	\toprule
	        & \multicolumn{5}{c|}{MAE\lob{} (meV)} & \multicolumn{4}{c}{EwT\hib{} (\%)}                                                                                                             \\
	Stastic & ID                                   & OOD Ads.                           & OOD Cat.    & OOD Both    & Avg.        & ID          & OOD Ads.   & OOD Cat.   & OOD Both   & Avg.       \\ \midrule
	Mean    & \rbs{378.4}                          & \rbs{469.1}                        & \rbs{397.8} & \rbs{407.3} & \rbs{413.1} & 10.98       & 5.54       & 9.43       & \rgd{6.04} & 8.00       \\
	Median  & \rgd{379.6}                          & \rgd{471.8}                        & \rgd{399.0} & \rgd{408.4} & \rgd{414.7} & \rgd{11.50} & \rgd{5.70} & \rgd{9.84} & \rbs{6.17} & \rgd{8.30} \\
	Mode    & 383.3                                & 475.5                              & 403.0       & 412.0       & 418.4       & \rbs{11.60} & \rbs{5.92} & \rbs{9.95} & 5.89       & \rbs{8.34} \\
	\bottomrule
\end{tabular}

%% file: tables/appendix/dist_acc.tex
\begin{tabular}{l|cc|cccc}
    \toprule
    Model                           & MAE (\r{A}) & RMSE (\r{A}) & EwT-0.2\r{A} (\%) & EwT-0.1\r{A} (\%) & EwT-0.05\r{A} (\%) & EwT-0.01\r{A} (\%) \\
    \midrule
    RDKit                           & 0.248       & 0.541        & 73.33             & 66.65             & 56.90              & 26.79              \\
    TGT-At-DP(no RDKit) + Refiner   & \rbs{0.152}       & \rbs{0.378}        & \rgd{80.10}             & \rgd{75.19}             & \rgd{70.38}              & \rbs{54.61}              \\
    TGT-At-DP(with RDKit) + Refiner & \rbs{0.152}       & \rbs{0.378}        & \rbs{80.53}             & \rbs{75.68}             & \rbs{70.80}              & \rgd{54.54}              \\
    \bottomrule
\end{tabular}